\documentclass[runningheads]{llncs}

\usepackage{eccv}

\usepackage{eccvabbrv}

\usepackage{graphicx}
\usepackage{booktabs}

\usepackage[accsupp]{axessibility}  

\usepackage[pagebackref,breaklinks,colorlinks,citecolor=eccvblue]{hyperref}

\usepackage[dvipsnames]{xcolor}
\usepackage{url}            
\usepackage{booktabs}       
\usepackage{amsfonts}       
\usepackage{nicefrac}       
\usepackage{microtype}      

\usepackage{graphicx}
\usepackage{comment}
\usepackage{amsmath,amssymb} 
\usepackage{xspace}
\usepackage[font=footnotesize,labelfont=bf]{caption} 
\usepackage{caption}
\usepackage{mathtools}
\usepackage{soul}
\usepackage{xspace}
\usepackage{enumitem}

\usepackage{bbding}
\usepackage{wasysym}
\usepackage{amssymb}
\usepackage{float}
\usepackage{dblfloatfix}
\usepackage{footnote}
\usepackage{array} 
\usepackage{dblfloatfix}
\usepackage{transparent}
\usepackage{xspace}
\usepackage{multirow}
\usepackage{amsfonts}
\usepackage{pifont}
\usepackage{algorithm}
\usepackage{listings}
\usepackage{bbold}
\usepackage{wrapfig}
\usepackage{subcaption}
\usepackage{sidecap}

\usepackage{etoolbox} 
\makeatletter
\AfterEndEnvironment{algorithm}{\let\@algcomment\relax}
\AtEndEnvironment{algorithm}{\kern2pt\hrule\relax\vskip3pt\@algcomment}
\let\@algcomment\relax
\newcommand\algcomment[1]{\def\@algcomment{\footnotesize#1}}
\renewcommand\fs@ruled{\def\@fs@cfont{\bfseries}\let\@fs@capt\floatc@ruled
  \def\@fs@pre{\hrule height.8pt depth0pt \kern2pt}%
  \def\@fs@post{}%
  \def\@fs@mid{\kern2pt\hrule\kern2pt}%
  \let\@fs@iftopcapt\iftrue}
\makeatother

\newcommand{\xmark}{\ding{55}}
\def\etal{\emph{et~al.}}
\def\eg{\emph{e.g.}}
\def\ie{\emph{i.e.}}
\def\etc{\emph{etc}}
\DeclareMathOperator*{\argmax}{arg\,max}

\newcolumntype{x}[1]{>{\centering\arraybackslash}p{#1pt}}
\newcolumntype{y}[1]{>{\raggedright\arraybackslash}p{#1pt}}
\newcolumntype{z}[1]{>{\raggedleft\arraybackslash}p{#1pt}}
\newlength\savewidth
\newcommand{\tablestyle}[2]{\setlength{\tabcolsep}{#1}\renewcommand{\arraystretch}{#2}\centering\footnotesize}

\definecolor{Highlight}{HTML}{39b54a}  
\definecolor{green}{HTML}{39b54a} 
\definecolor{red}{HTML}{cb4335} 

\DeclareCaptionLabelFormat{andtable}{#1~#2  \&  \tablename~\thetable}

\usepackage{orcidlink}

\begin{document}

\title{A Simple Latent Diffusion Approach for Panoptic Segmentation and Mask Inpainting} 
\titlerunning{A Simple Latent Diffusion approach for Panoptic Segmentation}
\author{Wouter Van Gansbeke\thanks{This work was done while the author was at Segments.ai, and the final training run was conducted at INSAIT. The author is now affiliated with Google DeepMind.} \and
Bert De Brabandere\inst{1}}
\authorrunning{W. Van Gansbeke and B. De Brabandere}
\institute{$^1$ Segments.ai}
\maketitle

\begin{abstract}
Panoptic and instance segmentation networks are often trained with specialized object detection modules, complex loss functions, and ad-hoc post-processing steps to manage the permutation-invariance of the instance masks. 
This work builds upon Stable Diffusion and proposes a latent diffusion approach for panoptic segmentation, resulting in a simple architecture that omits these complexities. Our training consists of two steps: (1) training a shallow autoencoder to project the segmentation masks to latent space; (2) training a diffusion model to allow image-conditioned sampling in latent space. This generative approach unlocks the exploration of mask completion or inpainting. 
The experimental validation on COCO and ADE20k yields strong segmentation results. Finally, we demonstrate our model's adaptability to multi-tasking by introducing learnable task embeddings. The code and models will be made available.\footnote{\href{https://github.com/segments-ai/latent-diffusion-segmentation}{https://github.com/segments-ai/latent-diffusion-segmentation}}

\end{abstract}

\section{Introduction}
The image segmentation task~\cite{long2015fully,minaee2021image} has gained popularity in
the literature, encompassing three popular subfields: semantic, instance, and panoptic segmentation. Over the years, segmentation tools have proven their usefulness for a wide range of applications, such as autonomous driving~\cite{cordts2016cityscapes}, medical imaging~\cite{menze2014multimodal},  agriculture~\cite{chiu2020agriculture}, and augmented reality~\cite{abu2018augmented,grauman2021ego4d}. Current methods are built upon convolutional networks~\cite{he2016deep} and transformers~\cite{vaswani2017attention,dosovitskiy2020image} to learn hierarchical image representations, and to simultaneously leverage large-scale datasets~\cite{zhou2019semantic,lin2014microsoft}. 
Earlier segmentation approaches relied on specialized architectures~\cite{he2017mask,cai2018cascade,tian2020conditional,wang2020solov2}, such as region proposal networks~\cite{ren2015faster} and dynamic convolutions~\cite{jia2016dynamic}. More recent approaches~\cite{cheng2021mask2former,carion2020end} advocate for an end-to-end strategy but introduce complex loss functions, \eg, bipartite matching. 
Some works have shown promising results without the necessity for 
labels~\cite{van2021revisiting,van2021unsupervised,van2022discovering,cho2021picie,wang2022freesolo,wang2023cut}, but likewise require highly-specialized modules, such as region proposal networks or clustering. Differently, we seek to leverage generative models to bypass the aforementioned components. 

\label{sec: introduction}
\begin{figure}[t]
    \centering
    \includegraphics[width=1.0\linewidth]{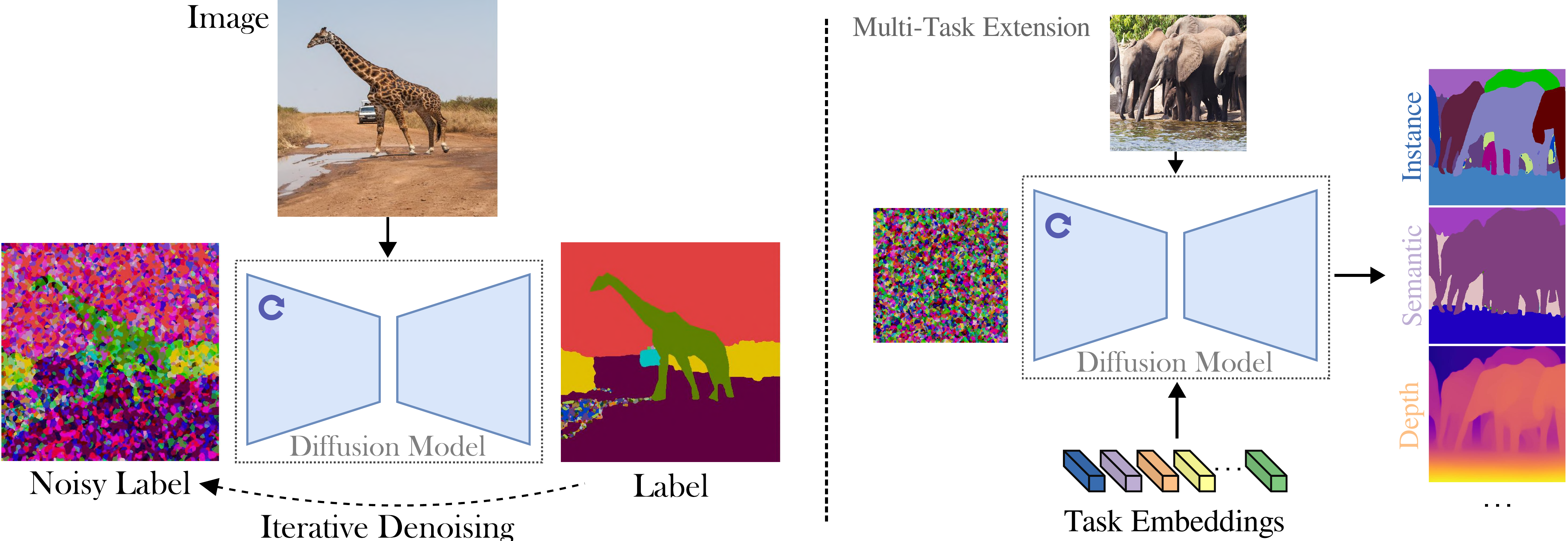}
    \caption{(\textit{Left:}) We present a simple generative approach for panoptic segmentation that builds upon Stable Diffusion~\cite{rombach2022high}. The key idea is to leverage the diffusion process to bypass complex detection modules and to unlock mask inpainting. The generative process is conditioned on RGB images to iteratively predict the masks. (\textit{Right:}) Our framework can be extended to a multi-task setting by introducing task embeddings.}
    \label{fig: teaser}
\end{figure}

\noindent This observation aligns with recent works~\cite{lu2022unified,chen2023generalist,kolesnikov2022uvim,chen2021pix2seq,chen2022unified} advocating for general computer vision models in an attempt to unify the field. These pioneering works limit the adoption of task-specific components but instead use a generative process.
In a similar vein, we take inspiration from recent text-to-image diffusion models~\cite{rombach2022high,ramesh2022hierarchical,nichol2021glide,dhariwal2021diffusion,ho2022classifier,ho2020denoising,ramesh2021zero, mizrahi2023} to tackle the segmentation task in a generative fashion. 
In addition to the diffusion model's general architecture, we further motivate this decision as follows:
(\textit{i}) diffusion models generate images with high photorealism and diversity; (\textit{ii}) perform on par with autoregressive priors while being more computationally efficient~\cite{song2021denoising,ramesh2022hierarchical}; (\textit{iii})  learn high-quality spatial representations, advantageous for dense prediction tasks;
(\textit{iv}) naturally exhibit image-editing capabilities. 
We now aim to build upon these properties.

\noindent To realize this objective, we introduce \textbf{LDMSeg}, a simple \textbf{L}atent \textbf{D}iffusion \textbf{M}odel for \textbf{Seg}mentation, visualized in Figure~\ref{fig: teaser}. Our contributions are fourfold:
\begin{itemize}[noitemsep, topsep=0pt]
    \item \textbf{Generative Framework:} We propose a fully generative approach based on Latent Diffusion Models (LDMs) for panoptic segmentation. We build upon Stable Diffusion~\cite{rombach2022high} to strive for simplicity and computational efficiency. 
    \item  \textbf{General-Purpose Design:} As a result, we circumvent specialized architectures, complex loss functions, and object detection modules, present in the majority of prevailing methods. Here, the denoising objective omits the necessity for object queries, region proposals, and Hungarian matching~\cite{kuhn1955hungarian} to handle the permutation-invariance of the instances.
    \item \textbf{Mask Inpainting:} We apply our method to scene-centric datasets. In contrast to prior art, we demonstrate mask inpainting for different sparsities.
    \item \textbf{Multi-Task Framework:} Our simple and general approach can easily be extended to train a single generative model for multiple tasks, like instance segmentation, semantic segmentation and depth prediction. Querying the model for a different task merely requires changing the task embedding. 
\end{itemize}
To the best of our knowledge, this paper presents the first latent diffusion approach that achieves strong results for panoptic segmentation, while also extending to mask inpainting and multi-task learning.
\section{Related Work}
\label{sec: related_work}

\textbf{Panoptic Segmentation.}
Panoptic segmentation~\cite{kirillov2019panoptic} has lately gained popularity as it combines semantic and instance segmentation. In particular, its goal is to detect and segment both \textit{stuff-like} (\textit{e.g.}, vegetation, sky, mountains, \textit{etc.}) and \textit{thing-like} (\textit{e.g.}, person, cat, car, \textit{etc.}) categories. 
Earlier works modified instance segmentation architectures to additionally handle (amorphous) \textit{stuff} categories as they are hard to capture with bounding boxes. For instance, Kirillov~\etal~showed promising results by making independent predictions using semantic and instance segmentation architectures~\cite{kirillov2019panoptic}, and later by integrating a semantic segmentation branch with a feature pyramid network (FPN) into Mask R-CNN~\cite{kirillov2019panopticfeat}.   
Other works~\cite{xiong2019upsnet,cheng2020panoptic} extend this idea by relying on specialized architectures and loss functions from both segmentation fields.  
More recent works~\cite{carion2020end,wang2021max,cheng2021mask2former,cheng2021maskformer,yu2022kmeans,zhang2021knet} handle \textit{things} and \textit{stuff} categories in a unified way via object queries and Hungarian matching~\cite{kuhn1955hungarian}. 
These works generally depend on specialized modules -- such as anchor boxes~\cite{ren2015faster}, non-maximum suppression~\cite{hosang2017learning,bodla2017soft}, merging heuristics~\cite{kirillov2019panoptic,xiong2019upsnet}, or bipartite matching algorithms~\cite{carion2020end,cheng2021mask2former,wang2021max}, \textit{etc.} -- to generate panoptic masks. Instead, we propose a task-agnostic generative framework to bypass these components. Consequently, we refrain from using task-specific augmentations, such as large-scale jittering or copy-paste augmentations~\cite{ghiasi2021simple}.
\vspace{0.07in}
\newline\textbf{General-purpose Frameworks.}
Similar to our work, a few task-agnostic solutions have been suggested that cast vision problems as a generative process. Kolesnikov~\etal~\cite{kolesnikov2022uvim} minimized task-specific knowledge by learning task-specific guiding codes with an LLM, to train a separate vision model. Chen~\etal~\cite{chen2022unified} simplified this procedure by framing vision tasks as language modeling tasks, within a single model. Lu~\etal~\cite{lu2022unified} also follow this route and show promising results for a large variety of vision and language tasks using a unified framework. Each work presents the input as a sequence of discrete tokens which are subsequently reconstructed via autoregressive modeling. Other works~\cite{wang2023images,bar2022visual,mizrahi2023} leverage masked image modeling to train a single model for multiple vision tasks. Differently, we leverage the denoising process in continuous latent space, which is well-suited to handle dense prediction tasks with high-dimensional inputs~\cite{ho2020denoising,chen2023generalist,rombach2022high}. 
\vspace{0.07in}
\newline\textbf{Denoising Diffusion Models.}
Denoising diffusion models~\cite{sohl2015deep,ho2020denoising,song2021denoising} were introduced as a new class of generative models. Recent strategies~\cite{dhariwal2021diffusion,nichol2021glide,ho2022classifier,ramesh2022hierarchical} additionally leverage text as guidance, \textit{e.g.}, via CLIP embeddings, to achieve results with impressive realism and control.
Building upon its success, a few diffusion-based solutions appeared in the segmentation literature. However, they have undesirable properties: (\textit{i}) the inability to differentiate between instances~\cite{amit2021segdiff,asiedu2022decoder,baranchuk2022abel,ji2023ddp,wang2024context}, (\textit{ii}) the necessity for specialized architectures and loss functions~\cite{gu2022diffusioninst,xu2023open}, or (\textit{iii}) the dependence on object detection weights and bit diffusion~\cite{chen2023generalist,chen2022analog}. 
The closest related work from Chen~\etal~\cite{chen2023generalist}, presents a framework for panoptic segmentation by leveraging the diffusion process in pixel-space.  
In contrast, we rigorously follow latent diffusion models by relying on continuous latent codes without the necessity for object detection data. Importantly, we can effortlessly leverage public image-diffusion weights~\cite{rombach2022high} as we keep the architecture task-agnostic.   

\section{Method}
\label{sec: method}
\begin{figure*}[t]
    \centering
    \includegraphics[width=1.0\linewidth]{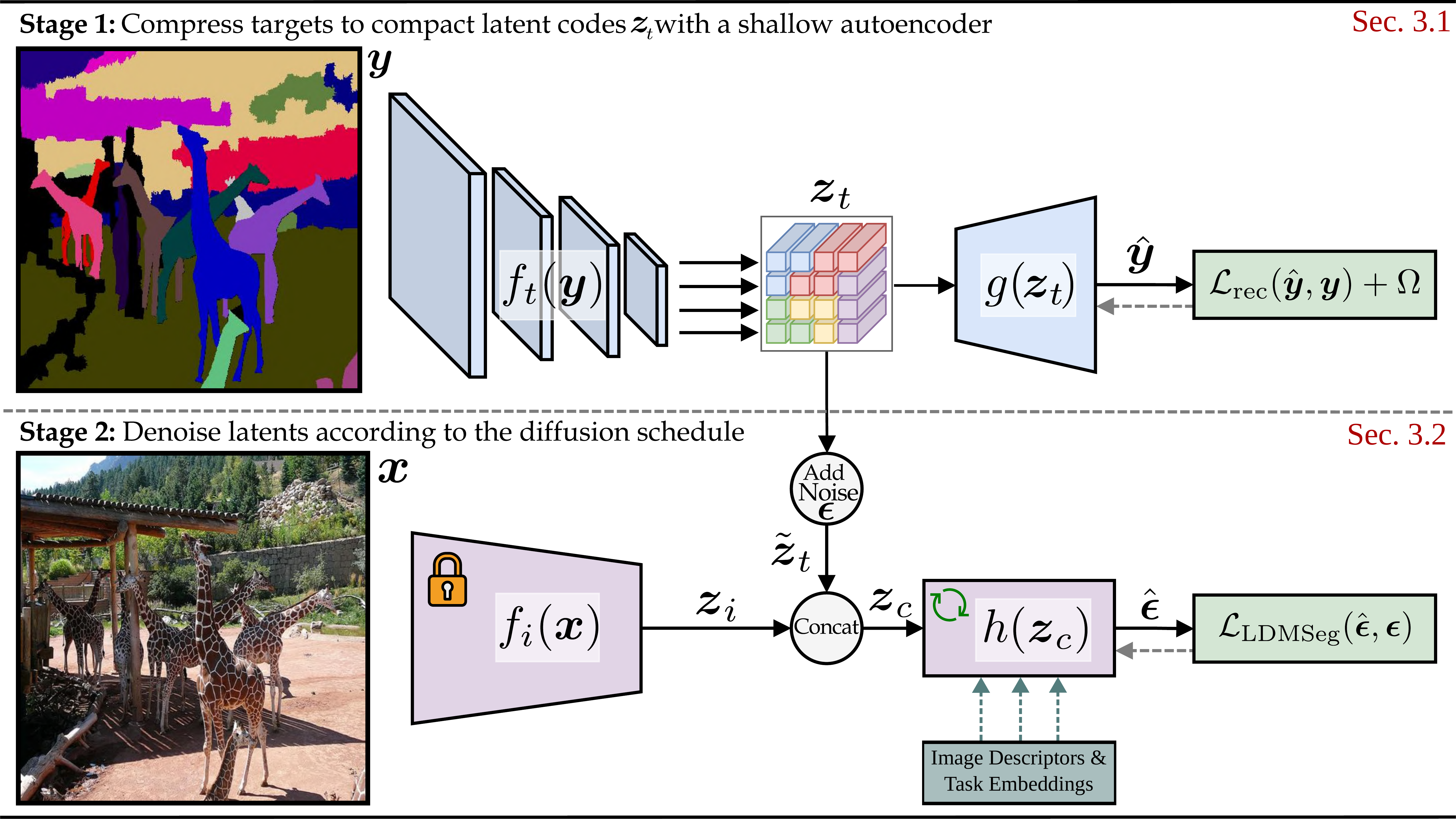}
    \caption{\textbf{Overview of LDMSeg.} Inspired by latent diffusion models, we present a simple diffusion framework for segmentation and mask inpainting. The approach consists of two stages: (\textit{i}) learn continuous codes $z_t$ with a shallow autoencoder on the labels (Sec.~\ref{subsec: stage1}); (\textit{ii}) learn a denoising function conditioned on image latents $z_i$ (Sec.~\ref{subsec: stage2}). In the second stage, the error between the predicted noise $\hat{\epsilon}$ and the applied Gaussian noise $\epsilon$ is minimized. During inference, we traverse the denoising process by starting from Gaussian noise. The models $f_t$ and $f_i$ respectively encode the labels and images. While we rely on the image encoder $f_i$ from Stable Diffusion~\cite{rombach2022high}, we focus on $f_t$ and $g$ for segmentation. We aim to prioritize generality by limiting task-specific components.}
    \label{fig: overview}
\end{figure*}

\textbf{Preliminaries.} 
This paper aims to train a fully generative model for panoptic segmentation via the denoising diffusion paradigm~\cite{ho2020denoising}. While this generative approach results in longer sampling times than discriminative methods, we justify this decision through four different lenses: \underline{(\textit{i}) ease of use} -- diffusion models omit specialized modules, such as non-maximum suppression or region proposal networks, are stable to train, and exhibit faster sampling than autoregressive models~\cite{van2016pixel};  \underline{(\textit{ii}) compositionality} -- this generative approach captures complex scene compositions with high realism and diversity, while also enabling image editing~\cite{ramesh2022hierarchical}; \underline{(\textit{iii}) dataset-agnostic} --  we rely on spatially structured representations that are not tied to predefined classes or taxonomies; \underline{(\textit{iv}) computational cost} -- latent diffusion models~\cite{rombach2022high} reduce the computational requirements by modeling latents instead of pixels with an autoencoder~\cite{kingma2014auto, van2017neural}. 
Motivated by these points, we create a framework that builds upon latent diffusion models' generative power~\cite{rombach2022high,ramesh2022hierarchical} for the segmentation task.
 
\vspace{0.07in}
\noindent\textbf{Problem Setup.} The considered segmentation task requires a dataset of images $\mathcal{X} = \{x_1, \ldots, x_n\}$ and corresponding ground truth panoptic segmentation masks $\mathcal{Y} = \{y_1, \ldots, y_n\}$. We don't make any distinction between \textit{things} or \textit{stuff} classes but handle all classes identically. 
We start by focusing on predicting panoptic IDs without the classes, rendering the method class-agnostic. Later, we extend our approach to also include class labels via task embeddings.

Assume that all images are of size $H \times W$. We train a segmentation model to realize the mapping  $\mathbb{R}^{3 \times H\times W} \rightarrow \mathbb{R}^{N \times H\times W}$. For each pixel the model performs a soft assignment over the instances $\{1, \ldots, N \}$.  
Let the latent representations $z_i$ and $z_t$ respectively refer to the image and target features after the projection to $D$-dimensional latents $\mathbb{R}^{D \times H/f \times W/f}$, where $f \in \mathbb{N}$ denotes the resizing factor. 

From a high-level perspective, our method has two key components: learning the prior over segmentation latents $z_t$ and learning a conditional diffusion process in latent space. First,  we train a shallow autoencoder to capture the prior distribution $p(z_t)$ that learns to compress the labels into compact latent codes $z_t$. Second, we train a diffusion process -- conditioned on the image and target features $p(y | z_t, z_i)$. This component is responsible for decoding noisy target features, guided by image features $z_i$. We make a similar derivation as~\cite{ramesh2022hierarchical} but condition the generative process on images. Formally, this two-step procedure allows us to construct the conditional distribution $p(y | x)$ via the chain rule as $p(y | x) = p(y | z_t, z_i) \cdot p(z_t)$.
Both terms, $p(z_t)$ and $p(y | z_t, z_i)$, are reflected in our framework as two separate training stages (see Figure~\ref{fig: overview}). In particular, Section~\ref{subsec: stage1} discusses the learning of the prior via autoencoding  and Section~\ref{subsec: stage2} leverages this prior to train a latent diffusion model. 

\subsection{Stage 1: Compress Targets}
\label{subsec: stage1}

In the first step, we train a network to compress the task-specific targets into latent codes. While we will focus on panoptic segmentation masks, we note that a similar strategy is applicable to other dense prediction targets (see Section~\ref{subsec: multi-task}). 

\vspace{0.07in}
\noindent\textbf{Motivation.}
The motivation to design a shallow autoencoder stems from the observation that segmentation maps differ fundamentally from images as they are lower in entropy. First, these masks typically contain only a small number of unique values as they only capture the object's general shape and location in the scene. Second, neighboring pixels are strongly correlated and often identical, resulting in largely spatially redundant information. We conclude that the segmentation task only necessitates a shallow network to efficiently compress the task's targets and to reliably capture the prior distribution $p(z_t)$. We hypothesize that this observation holds for a myriad of dense prediction tasks, such as panoptic segmentation, depth prediction, saliency estimation \textit{etc}. This also justifies why we refrain from using more advanced autoencoders that rely on computationally demanding architectures with adversarial or perceptual losses~\cite{johnson2016perceptual,zhang2018unreasonable}, \textit{e.g.}, VQGAN~\cite{esser2021taming}, typically used to encode images~\cite{rombach2022high}. 

\vspace{0.07in}
\noindent\textbf{Encoding.}
We analyze several encoding strategies to represent the input segmentation map $y$, latents $z_t$ and output $\hat{y}$. 
\newline
\underline{1. Input $y$}: Let $N$ denote the maximum number of instances per image. 
RGB-encoding (3 channels), bit-encoding ($\log_{2}N$ channels), one-hot-encoding ($N$ channels), or positional-encoding~\cite{vaswani2017attention} seem all justifiable options. However, one-hot encoding unnecessarily expands the input dimensions and bit encoding avoids the need for a fixed color palette for semantic segmentation. Hence, we follow~\cite{chen2023generalist} in representing the instances as bits.
\newline
\underline{2. Latent code $z_t$}: No vector quantization or dimensionality reduction is applied to the latent codes. 
We empirically found that directly using the continuous latents resulted in a simple architectural design with a small memory footprint (shallow) while simultaneously demonstrating encouraging segmentation results for both reconstruction and generation.
\newline
\underline{3. Output $\hat{y}$}: The output is one-hot encoded to train the autoencoder with a standard cross-entropy loss. In practice, we define $N$ output channels to reconstruct $N$ different instances in a scene, \textit{e.g.}, $128$.

\vspace{0.07in}
\noindent\textbf{Architectural Design.}
The architecture of the autoencoder follows a simple and shallow design. It comprises an encoder $f_t$ and decoder $g$, yielding the overall function $g \circ f_t$. The encoder $f_t$ includes only a few strided convolutions to compress the targets and is inspired by ControlNet~\cite{zhang2023adding}. For instance, targets with size 512$\times$512 can be resized efficiently to 64$\times$64, in order to leverage the latent space of Stable Diffusion~\cite{rombach2022high} by stacking 3 convolutions of stride 2. Similarly, the decoder $g$ consists of one or more transpose convolutions to upscale the masks and minimize a loss at pixel-level.  
As a consequence, the number of trainable parameters is at least 2 orders of magnitudes smaller than the amount of parameters in the diffusion model $h$ ($\approx 2$M \textit{vs.}~$800$M). The model's shallow design brings several advantages to the table: fast training, good generalization across datasets, and applicability to inpainting without architectural changes. 

\vspace{0.07in}
\noindent\textbf{Loss Function.}
The autoencoder aims to minimize the reconstruction error $\mathcal{L}_\text{rec}$ between the outputs $\hat{y}$ and the one-hot encoded segmentation masks $y$. While regression losses are a valid choice, we opt for a categorical loss due to the segmentation task's discrete nature. The reconstruction loss comprises two loss terms: (\textit{i}) the cross-entropy loss $\mathcal{L}_\text{ce}$ enforces unique and confident predictions for each pixel; (\textit{ii}) the mask loss  $\mathcal{L}_\text{m}$ further refines the segmentation masks by treating each instance individually, alleviating the need for exhaustively labeled images, in contrast to prior generalists~\cite{wang2023images,kolesnikov2022uvim}. This term is implemented via the BCE and Dice losses~\cite{sudre2017generalised,he2017mask}.
Note that this autoencoding strategy prevents the need for Hungarian matching~\cite{kuhn1955hungarian}. 

Typically, latent diffusion models incorporate a penalty term $\Omega$ into the loss formulation to align the bottleneck latents with a standard Gaussian distribution $\mathcal{N}(0, I)$. This term generally takes the form of a KL divergence, resulting in a variational autoencoder~\cite{kingma2014auto}. 
However, we empirically observed that weight decay regularization suffices to keep the weights $w$, and by extension the latents $z_t$, bounded (see Section~\ref{subsec: ablations}). Consequently, the magnitude of the model's weights $||w||_2$ is penalized, resulting in the final loss formulation:
\begin{equation}
\begin{split}
     \mathcal{L}_\text{AE}(w; y) 
     & = \mathcal{L}_\text{rec}(w; \hat{y}, y) + \Omega(z_t, w) \\
     & = \mathcal{L}_\text{ce}(w; \hat{y}, y) + \mathcal{L}_\text{m}(w; \hat{y}, y) + \lambda||w||_2^2,
\end{split}
\label{eq: loss_ae}
\end{equation}
where $\hat y$ denotes the reconstructed segmentation map.
Furthermore, we follow PointRend~\cite{kirillov2020pointrend} to select logits that correspond with uncertain regions.
This strategy limits the memory consumption as well as the total training time. 

\subsection{Stage 2: Train a Denoising Diffusion Model}
\label{subsec: stage2}

\begin{figure*}[t]
    \centering
    \includegraphics[width=1.0\linewidth]{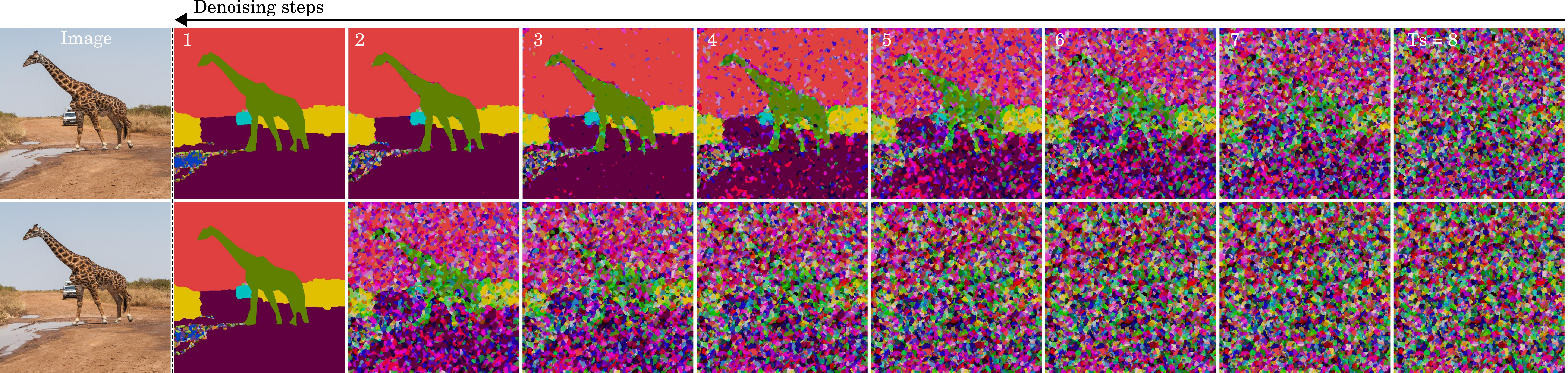}
    \caption{\textbf{Diffusion Process and SNR. } (\textbf{1}) \textbf{During training} we randomly sample a timestep from $[1, T]$ in the denoising process. We can increase the RGB-image's importance by strengthening the noise: (\textit{i}) Following~\cite{rombach2022high,chen2023generalist}, we downscale the latents $z_c$ using scaling factor $s \in \mathbb{R}$ and demonstrate its impact -- Row 1 ($s = 1.0$) is clearly easier to decode than row 2 ($s \approx 0.18$~\cite{rombach2022high}).  (\textit{ii}) Losses for timesteps near $0$ are further downscaled to avoid overfitting. Both strategies force the model to focus on the RGB image in generating plausible segmentation maps. Note, we don't apply explicit constraints to the prior distribution $p(z_t)$, \textit{e.g.}, match a standard Gaussian $\mathcal{N}(0, 1)$. (\textbf{2}) \textbf{During sampling} the denoising process is traversed from right to left in $T_s$ iterations.}
    \label{fig: denoising_process}
\end{figure*}

\noindent\textbf{Image-Conditioned Diffusion Process.}
\label{sec: diffussion_process}
The second stage of the framework models the function $h$ via a conditional diffusion process. We aim to learn a diffusion process over discrete timesteps by conditioning the model on images and noisy segmentation masks for each individual timestep.  We further follow the formulation of Stable Diffusion~\cite{rombach2022high}. This process is carried out in a joint latent space -- using our trained segmentation encoder $f_t$ and given image encoder $f_i$~\cite{rombach2022high} -- in order to project the targets and images to their respective latent representations $z_t$ and $z_i$. Next, we will discuss the training and inference procedures.

During training we linearly combine the noise $\epsilon$ with the latents $z_t$ for a randomly sampled timestep $j \in [1, T]$:
\begin{equation}
\label{eq: add_noise}
    \tilde{z}^j_t = \sqrt{\Bar{\alpha}_{j}}z_t + \sqrt{1 - \Bar{\alpha}_j}\epsilon,
\end{equation}
where $\Bar{\alpha}_j$ is defined by the noise schedule, following Rombach~\etal~\cite{rombach2022high}. 
The latents are subsequently fused $\{\tilde{z_t}^j, z_i\}$ via channel-wise concatenation as $z_c \in \mathbb{R}^{2D \times H/f \times W/f}$, before feeding them to the UNet~\cite{ronneberger2015u,rombach2022high} $h_\theta$, parameterized with weights $\theta$. Despite the examination of various fusing techniques (\textit{i.e.,} fusing intermediate features via cross-attention or via modality-specific branches) we found that straightforward concatenation at the input works surprisingly well.
At the output, the reconstruction error between the added Gaussian noise $\epsilon\sim\mathcal{N}(0, I)$ and the predicted noise $\hat\epsilon$ is minimized~\cite{ho2020denoising} as
\begin{equation}
\begin{split}
\label{eq: loss_ldmseg}
    \mathcal{L}_{\text{LDMSeg}}(\theta; \epsilon) 
    &= \mathbb{E}_{z_c, \epsilon\sim\mathcal{N}(0, I), j}\left[||\epsilon - h_\theta(z_c, j)||_2^2\right], 
\end{split}
\end{equation}
where each timestep $j$ is uniformly sampled from $[1, T]$. To reduce training time, we downscale the loss for small timesteps, \textit{i.e.} $j < 25\%\cdot T$. These latents have a high signal-to-noise ratio (SNR) and are thus relatively easy to denoise without modeling semantics. We refer to Algorithm~\ref{alg: forward_pass} for an overview of the forward pass. 

\begin{figure}[t]
\vspace{-0.2in}  
\noindent\begin{minipage}[t]{.50\linewidth} 
\centering
\begin{algorithm}[H]
\newcommand{\hlp}{\makebox[0pt][l]{\color{Mulberry!15}\rule[-3pt]{0.73\linewidth}{9pt}}}
\newcommand{\hlb}{\makebox[0pt][l]{\color{NavyBlue!15}\rule[-3pt]{0.73\linewidth}{9pt}}}
\algcomment{\fontsize{6.2pt}{0em}\selectfont \texttt{scheduler.weights}: array of length $T$ with loss weights}
\definecolor{codeblue}{rgb}{0.25,0.5,0.5}
\lstset{
  backgroundcolor=\color{white},
  basicstyle=\fontsize{7.2pt}{7.2pt}\ttfamily\selectfont,
  columns=fullflexible,
  breaklines=true,
  captionpos=b,
  commentstyle=\fontsize{7.2pt}{7.2pt}\color{codeblue},
  keywordstyle=\fontsize{7.2pt}{7.2pt},
}
\begin{lstlisting}[language=python, escapechar=@]
@\vspace{-2.0em}@
# f_i, f_t, h: encoders f_i & f_t, UNet h
# x: images of size [bs, 3, H, W]
# y: bit maps of size [bs, log(N), H, W]
# s, T: scaling factor, number of train steps
y = 2 * y - 1, 
x = x / 127.5 - 1   
z_t = f_t(y) * s 
z_i = f_i(x) * s 
j = torch.randint((bs,) 0, T)
noise = torch.randn_like(z_t)
# apply Eq. @\ref{eq: add_noise}@
z_t = scheduler.add_noise(z_t, noise, j)  
z_c = torch.cat([z_t, z_i], dim=1)
noise_pred = h(z_c, j)
loss = torch.sum((noise_pred - noise)**2, dim=[1,2,3])
loss = torch.mean(loss * scheduler.weights[j])
\end{lstlisting}
\vspace{-0.1in}
\captionof{algorithm}{Forward pass.}
\label{alg: forward_pass}
\end{algorithm}\vspace{-0.2in}
\end{minipage} %
\hspace*{\fill}
\begin{minipage}[t]{.48\linewidth}
\centering
\begin{algorithm}[H]
\newcommand{\hlp}{\makebox[0pt][l]{\color{Mulberry!15}\rule[-3pt]{0.73\linewidth}{9pt}}}
\newcommand{\hlb}{\makebox[0pt][l]{\color{NavyBlue!15}\rule[-3pt]{0.73\linewidth}{9pt}}}
\newcommand{\hlbb}{\makebox[0pt][l]{\color{NavyBlue!15}\rule[-3pt]{0.78\linewidth}{9pt}}}
\algcomment{
\fontsize{6.2pt}{0em}\selectfont \texttt{scheduler.inference\_steps}: chosen sampling timesteps \\
\fontsize{6.2pt}{0em}\selectfont \texttt{scheduler.step}: function - predict previous sample}
\definecolor{codeblue}{rgb}{0.25,0.5,0.5}
\lstset{
  backgroundcolor=\color{white},
  basicstyle=\fontsize{7.2pt}{7.2pt}\ttfamily\selectfont,
  columns=fullflexible,
  breaklines=true,
  captionpos=b,
  commentstyle=\fontsize{7.2pt}{7.2pt}\color{codeblue},
  keywordstyle=\fontsize{7.2pt}{7.2pt},
}
\begin{lstlisting}[language=python, escapechar=@]
@\vspace{-2.0em}@
# f_i, g: image and segmentation decoders
# h: denoising UNet
# x: image of size [3, H, W]
# T, Ts: #training steps, #inference steps
x = x / 127.5 - 1 
z_i = f_i(x) * s   @\hspace{0.01mm}@ 
z_t = torch.randn_like(z_i)  # Gaussian noise

for j in scheduler.inference_steps:
    z_c = torch.cat([z_t, z_i], dim=0)
    noise_pred = h(z_c, j)
    j_prev = j - T // Ts
    # apply Eq. @\ref{eq: ddim_step}@
    z_t = scheduler.step(z_t, noise_pred, j, j_prev) 
y_pred = g(z_t)  # decode latents
\end{lstlisting}
\vspace{-0.1in}
\captionof{algorithm}{Sampling process.}
\label{alg: sampling}
\end{algorithm}
\end{minipage}
\vspace{-0.15in}  
\end{figure}

During sampling, the denoising process is traversed from right to left (see Figure
\ref{fig: denoising_process}). It starts from Gaussian noise and progressively adds more details to the segmentation map as controlled by the input image. We can rely on the DDIM scheduler~\cite{song2021denoising} to apply this denoising process over a small number of sampling timesteps $T_s << T$. Here, the previous sample is computed as 
\begin{equation}
    \label{eq: ddim_step}
     \tilde{z}_t^{j-1} = \frac{\sqrt{\Bar{\alpha}_{j - 1}}}{\sqrt{\Bar{{\alpha}}_{j}}} (\tilde{z}_t^{j} - \sqrt{1 - \Bar{\alpha}_{j}}) \cdot \hat\epsilon + \sqrt{1 - \Bar{\alpha}_{j-1}} \cdot\hat\epsilon,
\end{equation}
which follows directly from Equation~\ref{eq: add_noise}.
Recall that this strategy allows us to model $p(y | z_t, z_i)$. Finally, Algorithm~\ref{alg: sampling} provides the details of the sampling process and Figure~\ref{fig: denoising_process} serves as an illustration.

\vspace{0.07in}
\noindent\textbf{Image Descriptors.} Complementary, we briefly experimented with adding guidance via Stable Diffusion's cross-attention layers to improve sample quality. This mechanism was initially intended for text embeddings, but can be used to feed any image descriptor. We didn't immediately observe improvements with self/weakly-supervised priors, like image embeddings from CLIP~\cite{radford2021learning}, or when using captions from BLIP~\cite{li2022blip}.
Hence, we can bypass the cross-attention layers with a skip connection in the class-agnostic setup. 

\vspace{0.07in}
\noindent\textbf{Task Embeddings.} To extend our framework to multiple tasks, we add learnable embeddings. We query the model for a certain task via its cross-attention layers. In particular, (class-aware) panoptic masks can be obtained by merging the semantic and instance predictions, which necessitates two task embeddings.  

\subsection{Segmentation Mask Inpainting}
\label{subsec: inpainting}

\noindent\textbf{Setup.} Our image-conditioned diffusion model is well-suited to complete partial segmentation masks. This is potentially useful for completing sparse segmentation masks obtained from projected point clouds or for interactive image labeling with rough brush strokes.  In contrast to image editing~\cite{rombach2022high}, the considered segmentation maps contain empty regions, which we initialize with $0$'s. We simulate different sparsities in Section~\ref{subsec: mask_inpainting}.
Existing state-of-the-art approaches are not designed for these applications as they decouple the output from the input via bipartite matching~\cite{cheng2021mask2former,carion2020end}. 
They require additional steps to match predictions with the given partial segmentation IDs (\textit{e.g.}, via majority voting). In contrast, diffusion models act on the corrupted masks by default.

\begin{wrapfigure}[20]{r}{0.5\textwidth}
\vspace{-0.6in}
\noindent\begin{minipage}[t]{1.0\linewidth} 
\centering
\begin{algorithm}[H]
\newcommand{\hlp}{\makebox[0pt][l]{\color{Mulberry!15}\rule[-3pt]{0.92\linewidth}{9pt}}}
\newcommand{\hlb}{\makebox[0pt][l]{\color{NavyBlue!15}\rule[-3pt]{0.92\linewidth}{9pt}}}
\newcommand{\hlbb}{\makebox[0pt][l]{\color{NavyBlue!15}\rule[-3pt]{0.98\linewidth}{9pt}}}
\definecolor{codeblue}{rgb}{0.25,0.5,0.5}
\lstset{
  backgroundcolor=\color{white},
  basicstyle=\fontsize{7.2pt}{7.2pt}\ttfamily\selectfont,
  columns=fullflexible,
  breaklines=true,
  captionpos=b,
  commentstyle=\fontsize{7.2pt}{7.2pt}\color{codeblue},
  keywordstyle=\fontsize{7.2pt}{7.2pt},
}
\begin{lstlisting}[language=python, escapechar=@]
@\vspace{-2.0em}@
# f_i, f_t: image encoder, segm. encoder
# g, h: segmentation decoder, denoising UNet
# x: image of size [3, H, W]
# y: sparse bit map of size [log(N), H, W]
# m: boolean mask w/ valid pixels, size [H, W]
# T, Ts: #train steps, #inference steps
y = 2 * y - 1 
@\hlbb@y[:, @$\sim$@ m] = 0  # set invalid regions to 0
x = x / 127.5 - 1 
z_i = f_i(x) * s  
@\hlbb@z_t_masked = f_t(y) * s 
@\hlbb@m = interpolate(m, size=z_i.shape[-2:])
z_t = torch.randn_like(z_i)
for j in scheduler.inference_steps:
    z_c = torch.cat([z_t, z_i], dim=0)
    noise_pred = h(z_c, j)
    # apply Eq. @\ref{eq: add_noise}@ (solve for z_t)
    @\hlb@z_t = scheduler.remove_noise(z_t, noise_pred, j) 
    @\hlb@z_t[:, m] = z_t_masked[:, m]  # keep latents
    j_prev = j - T // Ts
    # apply Eq. @\ref{eq: add_noise}@
    @\hlb@z_t = scheduler.add_noise(z_t, noise_pred, j_prev) 
y_pred = g(z_t)  # decode latents
\end{lstlisting}
\vspace{-0.1in}
\captionof{algorithm}{Inpainting process.}
\label{alg: inpainting}
\end{algorithm}
\end{minipage}
\end{wrapfigure}

\vspace{0.07in}
\noindent\textbf{Inpainting Process.}
Ideally, we can tackle inpainting problems out-of-the-box, \textit{i.e.,} without finetuning. To achieve this goal, the inference loop, previously  discussed in Algorithm~\ref{alg: sampling}, is modified.  Assume that we have a dataset of pairs $(y, m)$. Each pair contains a sparse segmentation mask $y \in \{0, 1\}^{\log_2 N\times H\times W}$, represented as a bit map, and a valid mask $m \in \{0, 1\}^{H\times W}$, represented as a boolean mask. Now, the diffusion process should fill in the missing regions in $y$, determined by the zeros in $m$. At each step of the denoising process, the latents corresponding to the given pixels in $m$ are fixed.
The key differences with the sampling process are highlighted in Algorithm~\ref{alg: inpainting}.

\section{Experiments}
\label{sec: experiments}

\begin{figure*}[t]
    \centering
    \includegraphics[width=1.0\linewidth]{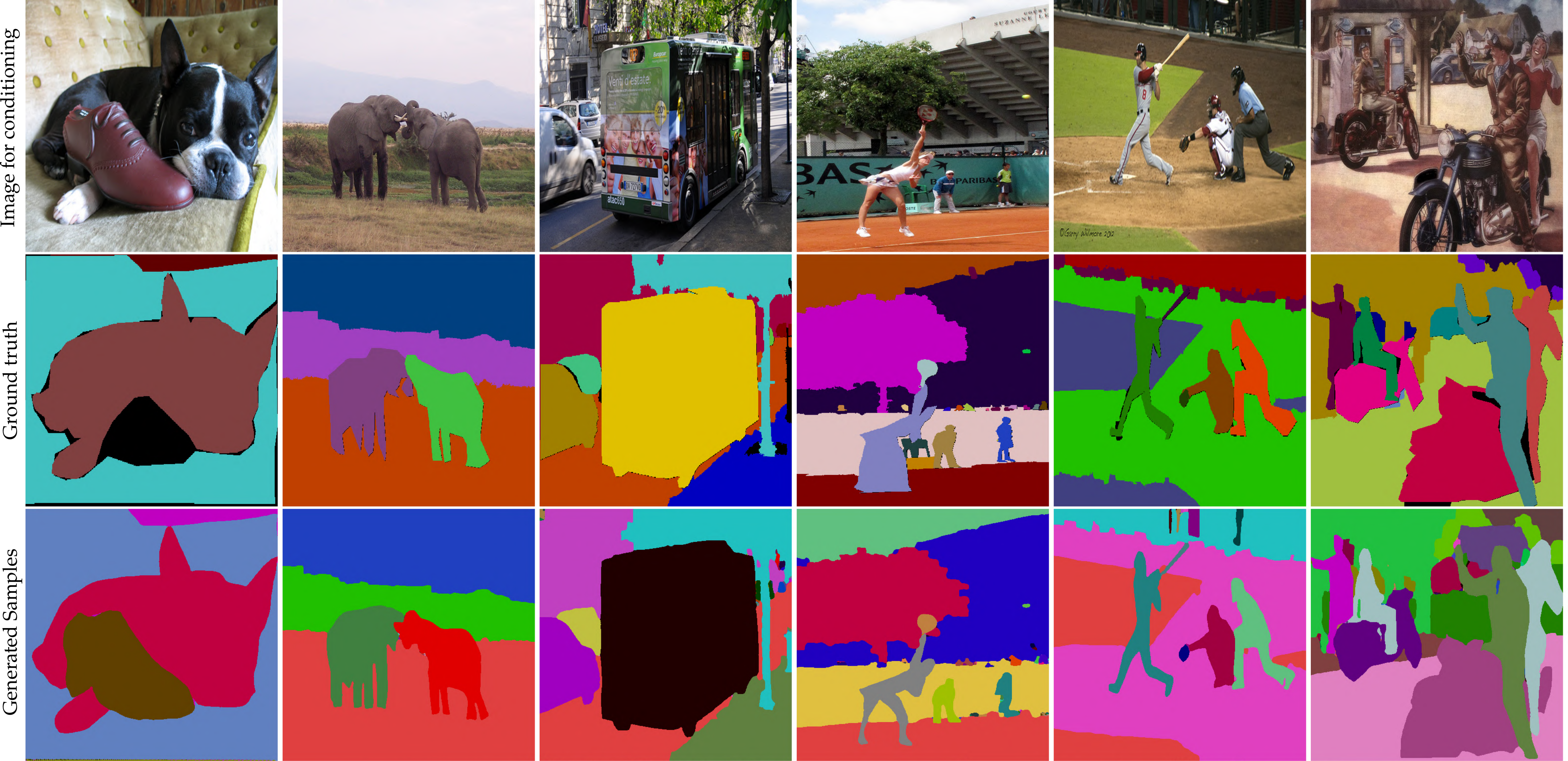}
    \caption{\textbf{Qualitative Results. } The figure displays results on COCO \texttt{val2017}. We follow the inference setup (Section~\ref{sec: diffussion_process}) to sample from our model. Only the $\argmax$ operator is applied for post-processing.  Our model disentangles overlapping instances in challenging scenes without complex modules or post-processing. To visualize, segments are assigned to random colors, and missing (VOID) pixels in the ground truth are black.}
    \label{fig: coco_results}
\end{figure*}

\begin{figure}[t]
    \centering
    \includegraphics[width=1.0\linewidth]{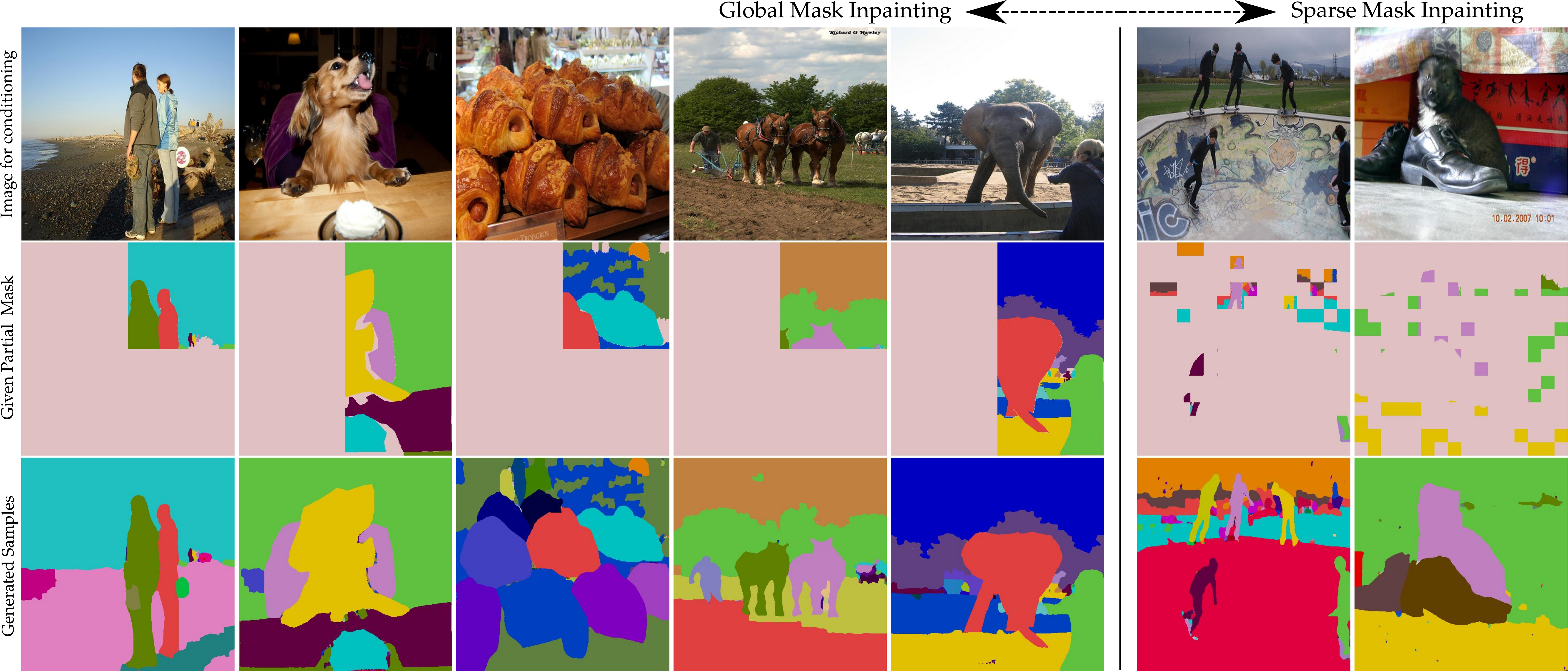}
    \caption{\textbf{Mask Inpainting.} The figure visualizes generated samples for different granularity levels by following Section~\ref{subsec: inpainting}. The model can fill in missing regions by propagating the partially given (random) segmentation IDs using an image-conditioned diffusion process.  Global mask inpainting (\textit{left}) results are reasonable out-of-the-box while sparse mask inpainting (\textit{right}) shows inaccuracies. We hypothesize that this can be addressed by further finetuning LDMseg on sparse inpainting masks. }
    \label{fig: inpainting}
\end{figure}

\textbf{Dataset.} We conduct the bulk of our experiments on COCO~\cite{lin2014microsoft} by relying on the panoptic masks with \textit{stuff} and \textit{things} classes. It contains 118k and 5k images for training and evaluation respectively. 

\vspace{0.07in}
\noindent\textbf{Architecture and Training Setup.}
The maximum number of detectable segments $N$ is set to $256$. 
We resize the input to $512 \times 512$, apply random horizontal flipping, and randomly assign integers from [$0, N-1$] to the segments.
The segmentation encoder $f_t$ processes the panoptic mask $y$ by leveraging 3 convolutional layers with stride 2 and SiLU~\cite{elfwing2018sigmoid} activations. This results in a resizing factor $f$ of $8$ and latents with size $4 \times 64 \times 64$. We rely on the image encoder $f_i$ from Rombach~\etal~\cite{rombach2022high} (VAE) to convert the image $x$ into latents. We adopt Stable Diffusion's pretrained weights to initialize the UNet $h$, and its rescaling factor $s$ to lower the SNR as $s \cdot z_c$~\cite{rombach2022high}. Notably, $4$ zero-initialized channels are appended to the first convolutional layer of $h$, allowing us to operate on the concatenated input $z_c$. Further, self-conditioning~\cite{chen2022analog} is used to improve sample quality.
Our segmentation decoder $g$ consists of 2 transpose convolutions, resulting in an upscaling factor of $4$. 
We randomly sample $j$ from $1000$ discrete timesteps and linearly decay the loss for the bottom $25$\% to lower the impact of samples with high SNR.
The AdamW~\cite{loshchilov2017decoupled} optimizer is adopted with a learning rate of $1e^{-4}$ and weight decay of $1e^{-1}$. Finally, the first stage is trained for 60k iterations with a batch size of $8$ while the second stage is trained for 50k iterations with a batch size of 256 unless stated otherwise.

\vspace{0.07in}
\noindent\textbf{Inference Setup and Evaluation Protocol.}
During inference, the DDIM scheduler~\cite{song2020improved} generates samples with $50$ equidistant timesteps in latent space. At the end of the denoising process, we decode and upscale the segmentation logits (output of $g$) with a factor of $2$ using bilinear interpolation. The $\argmax$ operator produces the final (discrete) per-pixel segmentation masks.
We benchmark our approach with the Panoptic Quality (PQ) evaluation protocol, defined by Kirillov~\etal~\cite{kirillov2019panoptic}. PQ is the product of two quality metrics: the segmentation quality which measures the intersection-over-union of matched segments (IoU) and the recognition quality which measures the precision and recall (F-score).

\subsection{Segmentation Results}

\textbf{Image-conditioned Mask Generation.}  
The model directly divides an image into non-overlapping semantic masks, capturing different objects in the scene, such as persons, horses and cars. Figure~\ref{fig: coco_results} showcases the predictions.

\begin{wrapfigure}[9]{r}{0.5\textwidth}
    \centering
    \includegraphics[width=1.0\linewidth]{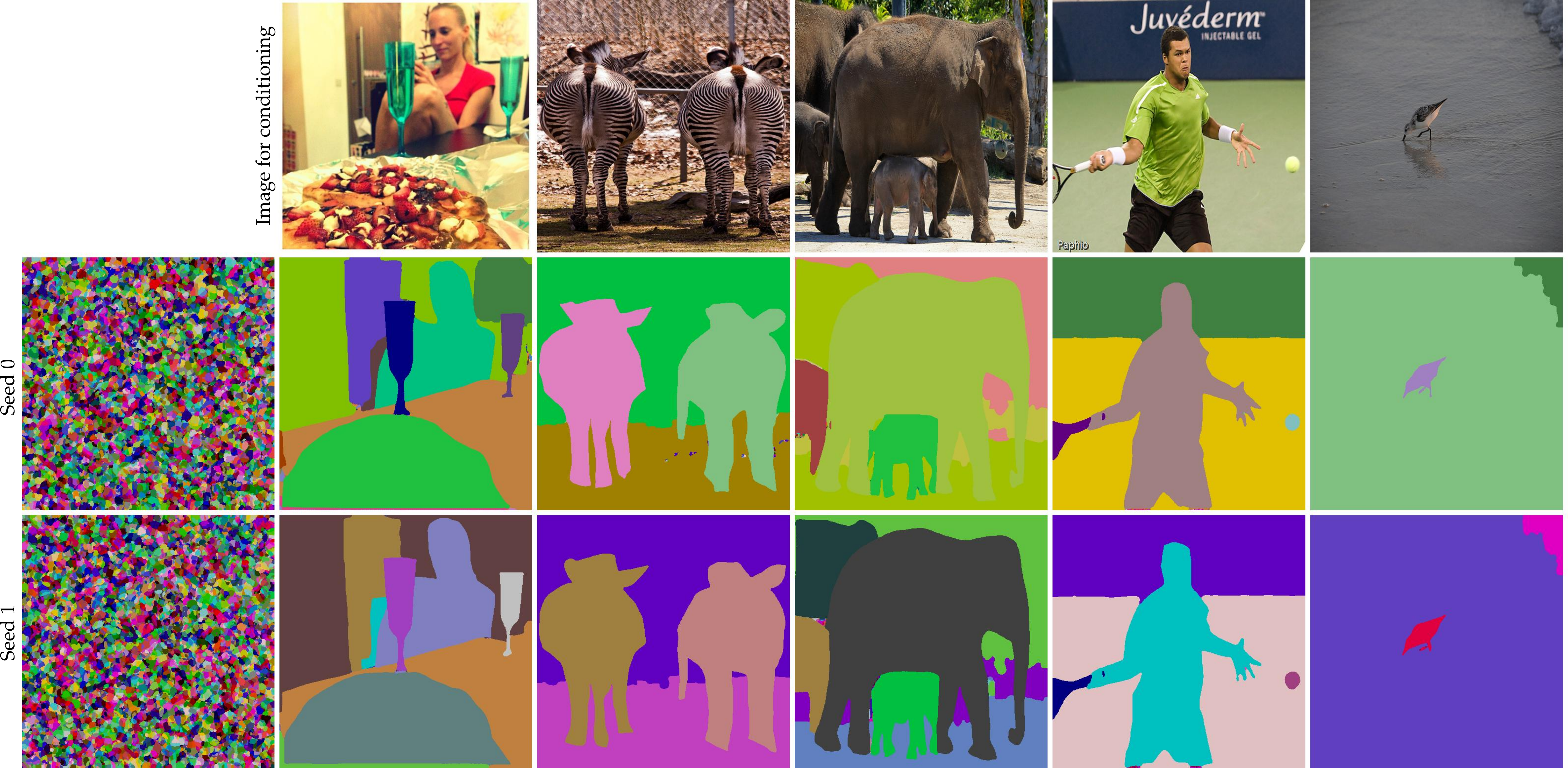}
    \vspace{-0.22in}
    \caption{\textbf{Impact of Gaussian Noise.}}
    \label{fig: noise_queries}
\end{wrapfigure}
\noindent As our model is inherently stochastic, we show the predictions when sampling different seeds in Figure~\ref{fig: noise_queries} (first column). 
Rows 2 and 3 show the predictions when starting
from the same noise map.
The key observation is that different maps generate different segmentation IDs. 
Compared to other frameworks, the sampled noise resembles the object queries in Mask2Former or DETR~\cite{cheng2021mask2former,carion2020end}, and the regions proposals in Mask R-CNN~\cite{he2017mask}. 
Table~\ref{tab: sota_comparison} quantitatively compares LDMSeg with prior art for the class-agnostic setup. We obtain 50.8\% PQ on COCO \texttt{val}. 
However, there is still a gap with specialized methods, \eg, Mask2Former~\cite{cheng2021mask2former} reaches 59.0\% PQ.  

\begin{figure*}[t]
\begin{minipage}[t]{.31\linewidth} 
\centering
\includegraphics[width=1.0\linewidth]{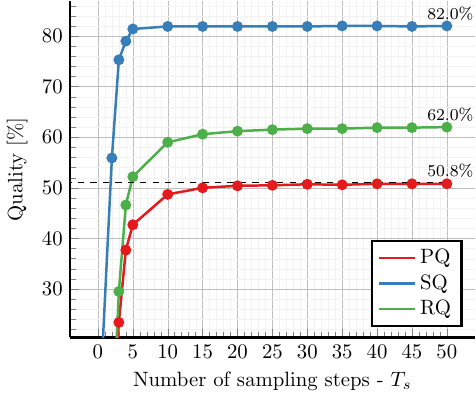}
\caption{\textbf{Sampling Steps.} The impact of sampling steps on PQ, SQ, and RQ.}
\label{fig: sampling_steps}
\end{minipage} %
\hspace{0.02\linewidth}
\begin{minipage}[t]{.31\linewidth} 
\centering
\includegraphics[width=1.0\linewidth]{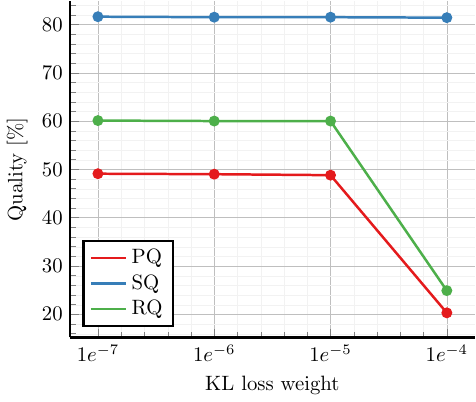}
\caption{\textbf{KL Loss.} The impact of a KL loss on PQ, SQ, and RQ.}
\label{fig: kl_ablation}
\end{minipage} %
\hspace{0.02\linewidth}
\begin{minipage}[t]{.31\linewidth} 
\centering
\includegraphics[width=1.0\linewidth]{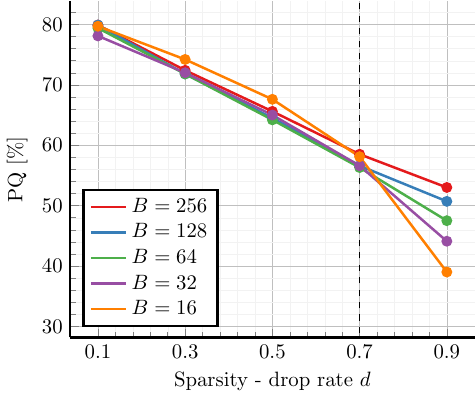}
\caption{\textbf{Mask Inpainting.} PQ for varying drop rates and block sizes.}
\label{fig: sparsity}
\end{minipage} %
\end{figure*}

\vspace{0.07in}
\noindent\textbf{Mask Inpainting.} 
\label{subsec: mask_inpainting}
Next, we mask random regions in the ground truth to simulate \textit{global} and \textit{local} completion tasks. 
Figure~\ref{fig: inpainting} shows the inpainting performance for two different granularity levels in mask $m$. The results demonstrate that LDMSeg is able to complete panoptic masks via the diffusion process, and without requiring additional components. 
Figure~\ref{fig: sparsity} displays the PQ metric for a wide range of drop rates $d$ and block sizes $B$. While our model is able to complete global regions, there is still room for improvement when considering sparse inpainting tasks (\eg, dropping $16\times 16$ pixels with a drop rate of $ 90\%$). For drop rates above $70\%$ and small block sizes, the reduction rate (negative slope) in PQ increases. Further improvements can be obtained by finetuning or modifying the autoencoder. For instance, shallower encoders $f_t$ could adapt better to sparse inputs, in exchange for lower reconstruction quality. Others~\cite{wang2023images,kolesnikov2022uvim,cheng2021mask2former} are not designed to complete partial masks without complex post-processing.

\definecolor{Highlight}{HTML}{39b54a}  
\renewcommand{\hl}[1]{\textcolor{Highlight}{#1}}
\newcommand{\hlbb}{\makebox[0pt][l]{\color{NavyBlue!15}\rule[-3pt]{0.8\linewidth}{11pt}}}
\begin{table}[t]
\caption{\textbf{Class-agnostic Comparison.} We compare with Mask2Former on COCO \texttt{val2017}. Training details for Mask2Former are specified in Supplement {\color{red}A}, following SAM~\cite{kirillov2023segment} with MAE~\cite{he2022masked} or DINOv2~\cite{oquab2023dinov2} ($\star$) initialization.  ($\dagger$) denotes the adoption of Stable Diffusion's VAE with 84M parameters, resulting in $40\times$ more parameters (84M \vs 2M) for the same performance (Section~\ref{subsec: stage1}). $PQ_{inpaint}$ is the averaged PQ metric when dropping $16\times 16$ pixels with probabilities from 10 to 90\%.}
\tablestyle{10.0pt}{1.0}
    \begin{tabular}{l | c | c | c}
    \toprule
    \textbf{~~Method} & \textbf{Backbone} & \textbf{PQ} $\uparrow$ & \textbf{PQ}$_{inpaint}$ $\uparrow$ \\
     \hline
     \textit{Specialist approaches:} & & & \\
    ~~MaskFormer~\cite{cheng2021maskformer} & ViT~\cite{dosovitskiy2020image}  & 54.1 & \xmark \\
    ~~Mask2Former~\cite{cheng2021mask2former} & ViT~\cite{dosovitskiy2020image} & 56.5 & \xmark \\
    ~~Mask2Former$^{\star}$~\cite{cheng2021mask2former} & ViT~\cite{dosovitskiy2020image} & 59.0 & \xmark \\
    \hline
    \textit{Generalist approaches:} & & & \\
    ~~\textbf{LDMSeg}$^\dagger$ & UNet~\cite{ronneberger2015u} & \textbf{50.9}  & -- \\ 
    \hlbb ~~\textbf{LDMSeg} & UNet~\cite{ronneberger2015u} & 50.8  & \textbf{61.3} \\
    \bottomrule
    \end{tabular}
\label{tab: sota_comparison}
\end{table}

\renewcommand{\hlbb}{\makebox[0pt][l]{\color{NavyBlue!15}\rule[-3pt]{0.91\linewidth}{11pt}}}
\begin{table}[t]
\caption{\textbf{State-of-the-art Comparison.} The table presents the panoptic and semantic segmentation results on COCO \texttt{val2017} and ADE20k \texttt{val} respectively.}
\resizebox{1.0\linewidth}{!}{
\tablestyle{10.0pt}{1.0}
    \begin{tabular}{l | c | c | c }
    \toprule
    &      &  \textbf{COCO}~\cite{lin2014microsoft} & \textbf{ADE20k}~\cite{zhou2019semantic} \\
    \textbf{~~Method} & \textbf{Backbone} & \textbf{PQ} $\uparrow$  & \textbf{mIoU} $\uparrow$ \\
    \hline
    \textit{Specialist approaches:} & & \\
    ~~PanopticFPN~\cite{kirillov2019panopticfeat} & ResNet~\cite{he2016deep} & 44.1 & -- \\
    ~~DETR~\cite{carion2020end} & ResNet~\cite{he2016deep} & 45.6 & -- \\
    ~~MaskFormer~\cite{cheng2021maskformer} & ResNet~\cite{he2016deep} & 46.5 & 44.5 \\
    
    ~~UPerNet~\cite{xiao2018unified} & Swin-L~\cite{liu2021swin} & \xmark & 52.1 \\
    ~~DDP~\cite{ji2023ddp} & Swin-L~\cite{liu2021swin} & \xmark & 53.2 \\
~~Mask2Former~\cite{cheng2021mask2former} & Swin-L~\cite{liu2021swin} & \textbf{57.8} & \textbf{56.1} \\
    \hline
    \textit{Generalist approaches:} & & \\
    ~~Painter~\cite{wang2023images} & ViT~\cite{dosovitskiy2020image} & 41.3 & 47.3 \\
    ~~UViM~\cite{kolesnikov2022uvim} & ViT~\cite{dosovitskiy2020image} & \hspace{0.37in}43.1 (45.8) & -- \\
    \hlbb ~~\textbf{LDMSeg} & UNet~\cite{ronneberger2015u} & \textbf{44.3} & \textbf{52.2} \\ 
    \bottomrule
\end{tabular}
}
\label{tab: sota_pq} 
\end{table}

\begin{table}[t]
\caption{\textbf{Multi-Task Performance.} The table reports results for semantic segmentation, panoptic segmentation and relative depth (between 0 - 1) on COCO \texttt{val2017}.}
\resizebox{1.0\linewidth}{!}{
\tablestyle{8.0pt}{1.0}
    \begin{tabular}{l c  c c c}
    \toprule
    & & \textbf{Semantic Seg.} & \textbf{Panoptic Seg.} & \textbf{Relative Depth} \\
    \cmidrule{3-3} \cmidrule{4-4} \cmidrule{5-5}
    \textbf{Setting} & \textbf{\# Iters}  & \textbf{mIoU} $\uparrow$ & \textbf{PQ} $\uparrow$ & \textbf{RMSE} $\downarrow$  \\
    \hline
    Multi-Task & 50 & 60.1   & 44.1 & 0.075 \\
    Single-Task & 50 & 61.5  & 44.3 & 0.067 \\   
    \bottomrule
    \end{tabular}
}
\label{tab: multi_task}
\end{table}

\vspace{0.07in}
\noindent\textbf{State-of-the-art Comparison.} 
As the panoptic task combines instance and semantic segmentation, we introduce two learnable task embeddings. Additionally, we adopt a ViT-B~\cite{dosovitskiy2020image} as the image encoder to improve the performance (see Supplement {\color{red}{B}} for a component analysis). 
Table~\ref{tab: sota_pq} compares LDMSeg with state-of-the-art methods after training for 100k iterations:
\begin{itemize}[noitemsep, topsep=0pt]
\item \textit{LDMSeg \vs Specialists}: Our generative approach is competitive with several specialized approaches~\cite{xiong2019upsnet,kirillov2019panopticfeat,carion2020end}. For instance, we can match the performance of PanopticFPN~\cite{kirillov2019panoptic} while not requiring region proposals. Instead, we rely on the diffusion process to solve the permutation-invariance of the instances. 

\item \textit{LDMSeg \vs Generalists}:
(\textit{i}) Painter necessitates two separate encoding schemes for \textit{things} and \textit{stuff} categories, as well as non-maximum-suppression (NMS). LDMSeg outperforms Painter ($44.3\%$ \vs $41.3\%$) while not relying on these components, nor on complex merging strategies. This renders our approach more general. In fact, our post-processing time is an order of magnitude smaller (see Supplement {\color{red}B}). (\textit{ii}) Additionally, LDMSeg is competitive with UViM's~\cite{kolesnikov2022uvim} public model ($44.3\%$ \vs $43.1\%$). We hypothesize that the disparities in UViM's results could be attributed to its reliance on specific code lengths and code dropout~\cite{kolesnikov2022uvim}, in order to train its autoregressive language model (LM). In contrast, we leverage latent codes centered around zero using a shallow autoencoder, enabling better control. To increase the denoising difficulty, we simply lower the scaling factor (see Section~\ref{subsec: stage2}). 
Pix2Seq-$\mathcal{D}$~\cite{chen2023generalist} relies on a ResNet backbone that is pretrained with additional bounding-box annotations (Objects365~\cite{shao2019objects365}) and achieves $50.3\%$ PQ.  However, its unreleased weights poses a challenge in adopting this paradigm. To our knowledge, LDMSeg is first in demonstrating that a latent diffusion process can bypass object detection data and its related modules. Larger datasets and a higher resolution will help in closing the gap.

\end{itemize}
Finally, we tackle semantic segmentation and show results for ADE20k~\cite{zhou2019semantic}.
This dataset contains 20k training images and 2k validation images, covering 150 semantic classes. LDMSeg surpasses well-performing methods~\cite{wang2023images,cheng2021maskformer,xiao2018unified}, reaching 52.2\% mIoU on the validation set. Interestingly, our model is able to capture the mapping from latents to the respective classes without making changes to its design. While DDP~\cite{ji2023ddp} performs slightly better (52.2\% \vs 53.2\%), it can not handle the permutation-invariance of the instance masks. This limitation stems from relying on the diffusion process as a refinement step and its necessity for a direct mapping between inputs and outputs (\eg, in semantic segmentation). 

\subsection{Multi-Task Learning Results} 
\label{subsec: multi-task}
 We now broaden the scope and extend our approach to a multi-task setting. Table~\ref{tab: multi_task} reports the results when training on 3 vision tasks for 100k iterations on COCO in total. We handle all tasks identically, including the same scaling factor and augmentations.
Given that the single-task setup has been trained for 3 $\times$ longer, it serves as an upperbound (\eg, 44.2\% \vs 44.3\% for PQ). Most importantly, we conclude that LDMSeg can be trained on multiple dense prediction tasks simultaneously by leveraging task embeddings. We refer to Supplement {\color{red}{A}} for more details and visualizations.

\subsection{Ablation Studies}
\label{subsec: ablations}
We ablate LDMSeg in the class-agnostic setting after training for 50k iterations.
\begin{itemize}[noitemsep, topsep=0pt]
    \item The impact of the number of inference steps is shown in Figure~\ref{fig: sampling_steps}. We observe that the PQ starts plateauing at 20 iterations.  Longer inference schedules can further improve the recognition quality (a reduction in false positives and negatives), while keeping the segmentation quality constant ($\sim 80\%$). 
    \item Figure~\ref{fig: kl_ablation} demonstrates the impact of adding a KL loss to the autoencoding objective in Equation~\ref{eq: loss_ae}. While it aligns the latents with a standard Gaussian distribution, increasing its loss weight beyond $1e^{-5}$ hurts the reconstruction quality. We conclude that weight decay suffices to keep the latents bounded.
    \item Table~\ref{tab: sota_comparison} verifies the hypothesis that panoptic masks don't require a powerful VAE. In particular, we finetuned Stable Diffusion's VAE~\cite{rombach2022high} instead of our shallow autoencoder (see Section~\ref{subsec: stage1}). However, this results in a similar PQ (50.9\% \vs 50.8\%). Our shallow autoencoder contains $40\times$ less parameters, which reduces the training time of stage 2 up to 20\%  (see Section~\ref{subsec: stage2}). 
\end{itemize}  
\section{Conclusion}
\label{sec: conclusion}
We presented LDMSeg, a simple yet powerful latent diffusion approach for panoptic segmentation and mask inpainting. In contrast to prior art, we leverage plain latent diffusion models by building upon Stable Diffusion~\cite{rombach2022high}. In summary, the proposed image-conditioned diffusion process has the following advantages: (\textit{i}) it bypasses specialized modules, such as region proposals and bipartite matching; (\textit{ii}) our model unlocks sparse panoptic mask completion without finetuning; (\textit{iii}) the approach can easily be extended to a multi-task setting by introducing task embeddings. 
The experiments show that LDMSeg is versatile while also outperforming the majority of prior \textit{generalists}. 
Due to its simple and general design, we believe there is still room for improvement in terms of accuracy and sampling speed. 
Evident future directions include: training LDMSeg on larger datasets and incorporating more dense prediction tasks (\eg, edge detection).
Hence, we hope that this work will spark further interest in designing general-purpose approaches for dense prediction tasks.

\newpage
\appendix
\setcounter{section}{0}
\renewcommand{\thesection}{\Alph{section}}
\setcounter{figure}{0}
\setcounter{table}{0}
\renewcommand{\thefigure}{S\arabic{figure}}
\renewcommand{\thetable}{S\arabic{table}}

\definecolor{rowcolor}{RGB}{209,154,128}

 We discuss the implementation details in Supplement~\ref{sec: implementation_suppl}, additional results in Supplement~\ref{sec: results_suppl}, limitations in Supplement~\ref{sec: limitations} and the broader impact in Supplement~\ref{sec: broader_impact}.

\section{Implementation Details}
\label{sec: implementation_suppl}
\paragraph{Model Card.} Our best model is trained for 100k iterations on COCO with mixed precision training on 8 $\times$ 40GB NVIDIA A100 GPUs using Google Cloud. We rely on the pretrained Stable Diffusion~\cite{rombach2022high} weights provided by Hugging Face~\cite{huggingface2023stablediffusionv14}. We also adopt its settings for the noise scheduler. The code is developed in Pytorch~\cite{pytorch} and will be made available as well as our models.

\begin{figure*}
  \centering
   \includegraphics[width=1.0\linewidth]{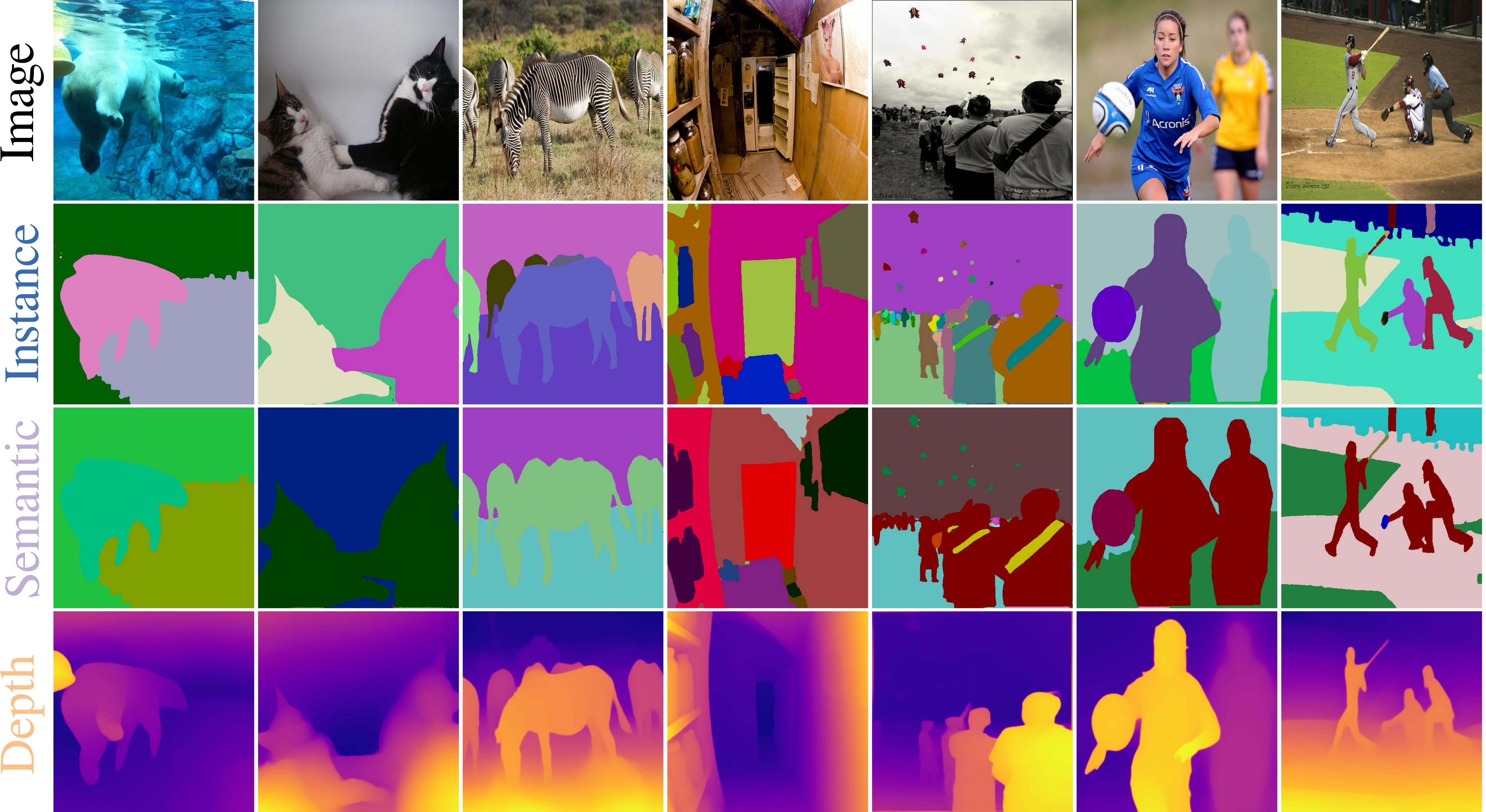}
   \caption{\textbf{Multi-Task Setup - Qualitative Results. } The figure displays the results for several images in the COCO val set. We can query the model for multiple tasks as it has learned their respective task embeddings.}
   \label{fig: multi-task}
\vspace{-0.15in}
\end{figure*}

\paragraph{Multi-Task Setup.} This section provides additional information on the multi-task extension for dense prediction, with minor adaptations. Consider the three fundamental vision tasks: instance segmentation, semantic segmentation and depth prediction.
The instance and semantic tasks both utilize the same shallow autoencoder to generate continuous latent codes.
Similarly, to compress the depth maps, we rely on the same shallow autoencoder architecture as its segmentation counterpart. We only change the input and output channels to one channel. As COCO does not contain depth annotations, we rely on the predictions from MiDaS~\cite{birkl2023midas} to obtain pseudo ground truth. Note that this model predicts relative depth. All tasks use the same set of augmentations and scaling factors, as discussed in the main paper. 
To enable multi-tasking, we introduce learnable task embedding (786-dimensional) via the cross-attention layers of the UNet. This allows us to query the model for a specific task. Figure~\ref{fig: multi-task} visualizes the results for each task by only changing the task embedding. We observe that the model can predict accurate instance, semantic and depth maps for a given image. Finally,  
given our shallower encoder and task embeddings, a comparison with Marigold~\cite{ke2024repurposing}, a concurrent work on depth estimation, could be insightful.

\paragraph{Mask2Former Baselines.}
Mask2Former~\cite{cheng2021mask2former} is a specialized segmentation framework that produces excellent results for panoptic segmentation. We follow the training recipe from ViTDet~\cite{li2022exploring} and SAM~\cite{kirillov2023segment} to leverage plain ViT backbones~\cite{dosovitskiy2020image} with MAE pretrained weights~\cite{he2022masked}. Specifically, the model consists of the vision transformer backbone, a shallow neck, and a mask decoder. The latter contains $6$ masked attention decoder layers and $128$ object queries, following~\cite{cheng2021mask2former}. The loss requires Hungarian matching~\cite{kuhn1955hungarian} to handle the permutation invariance of the predictions during training. To report the results, we follow its post-processing strategy to combine the classification and mask branches. We adopt the same augmentations as in the main paper, \ie, square resizing and random horizontal flipping. This baseline strikes a good balance between performance, complexity, and training speed. 
Additionally, we provide results by relying on the backbone and pretrained weights of DINOv2~\cite{oquab2023dinov2}, as we found this to outperform MAE pretrained weights for a ViT-B backbone. We train the models with a batch size of $32$ and a learning rate of $1.5e^{-4}$ for $50$k iterations on 8 $\times$ 16GB V100 GPUs. 

\paragraph{Evaluation Procedure.}
Our model produces excellent predictions when only relying on the $\argmax$ operator. No additional processing is used for the visualizations (see row 3 in figures~\ref{fig: generation_v1} and~\ref{fig: generation_v2}). To report the final PQ metric, however, we eliminate noise by thresholding the predictions at $0.5$ (after applying softmax) and filtering out segments with an area smaller than $512$.  These results are shown in the last row of figures~\ref{fig: generation_v1} and~\ref{fig: generation_v2}. Notice that Mask2Former's training objective does not impose exclusive pixel assignments, hence it needs additional post-processing steps.

\paragraph{Simple Post-Processing.} Panoptic segmentation combines instance and semantic segmentation. After efficiently decoding the latents, we will obtain the panoptic mask by starting from the instances. We subsequently take a majority vote using the predicted semantic mask for each instance. We carry out the following steps and refer to Supplement~\ref{sec: results_suppl} for more information on the inference time: 
\definecolor{codegreen}{rgb}{0,0.6,0}
\definecolor{codegray}{rgb}{0.5,0.5,0.5}
\definecolor{codepurple}{rgb}{0.58,0,0.82}
\definecolor{backcolour}{rgb}{0.95,0.95,0.92}

\lstdefinestyle{mystyle}{
    backgroundcolor=\color{backcolour},   
    commentstyle=\color{codegreen},
    keywordstyle=\color{magenta},
    numberstyle=\tiny\color{codegray},
    stringstyle=\color{codepurple},
    basicstyle=\ttfamily\footnotesize,
    breakatwhitespace=false,         
    breaklines=true,                 
    captionpos=b,                    
    keepspaces=true,                 
    numbers=left,                    
    numbersep=5pt,                  
    showspaces=false,                
    showstringspaces=false,
    showtabs=false,                  
    tabsize=2
}

\lstset{style=mystyle}
\begin{lstlisting}[language=Python]
def postprocess_panoptic(mask_logits, semantic_logits):
    """
    Convert predictions to panoptic masks.

    Inputs:
        mask_logits: np.array of size [N, H, W]
        semantic_logits: np.array of size [N, H, W]
    Outputs:
        panoptic_seg: np.array of size [H, W]
        segments_to_categories: dict
    """

    panoptic_seg = np.argmax(mask_logits, axis=0)
    semantic_seg = np.argmax(class_logits, axis=0)
    
    segments_ids = {}
    for segment_id in np.unique(panoptic_seg): 
        instance_mask = panoptic_seg == segment_id
        if not_confident_or_small(instance_mask):
            panoptic_seg[instance_mask] = VOID_id
            continue
        counts = np.bincount(semantic_seg[instance_mask]])
        class_id = np.argmax(counts)
        segments_to_categories[segments_id] = class_id
        
    return panoptic_seg, segments_ids
\end{lstlisting}

\section{Additional Results}
\label{sec: results_suppl}
\begin{figure*}
  \centering
   \includegraphics[width=1.0\linewidth]{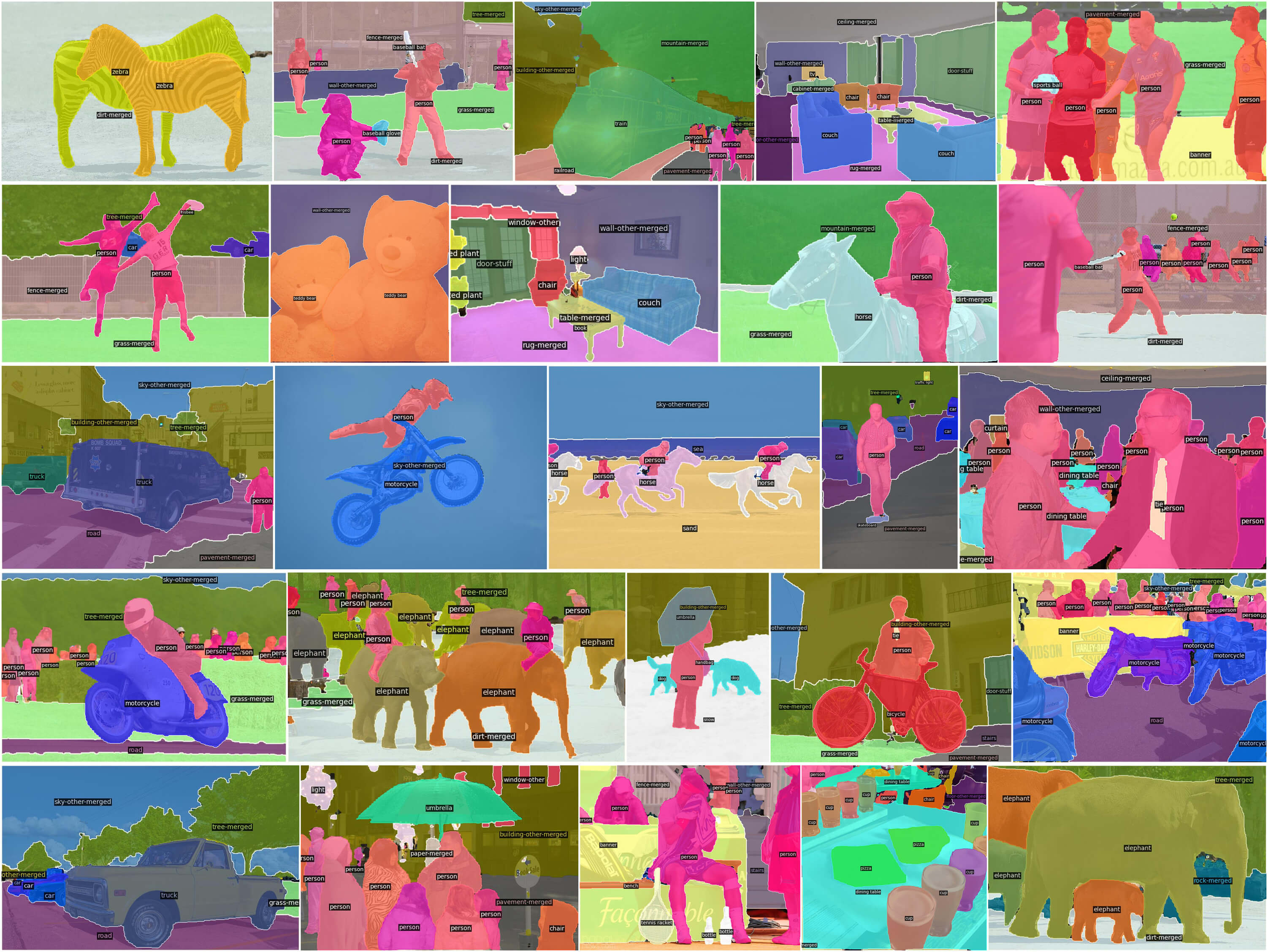}
   \caption{\textbf{Panoptic Segmentation - Qualitative Results. } The figure displays the panoptic segmentation for several images in the COCO val set.}
   \label{fig: pq_results}
\vspace{-0.15in}
\end{figure*}

\begin{figure*}
  \centering
   \includegraphics[width=0.92\linewidth]{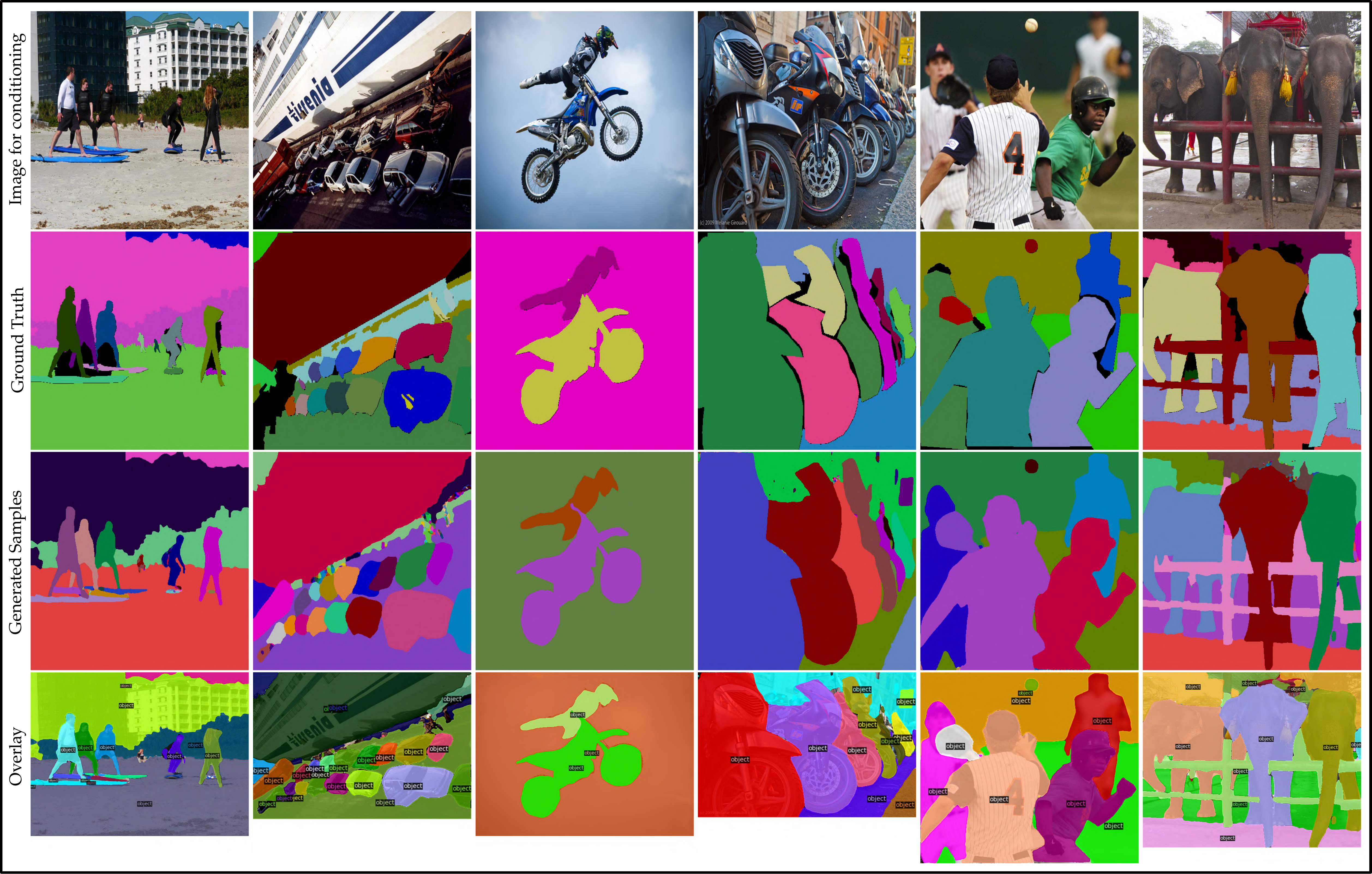}
   \caption{\textbf{Examples on COCO (1). } The figure displays the generated masks on the COCO val set.}
   \label{fig: generation_v1}
\end{figure*}

\begin{figure*}
  \centering
   \includegraphics[width=0.92\linewidth]{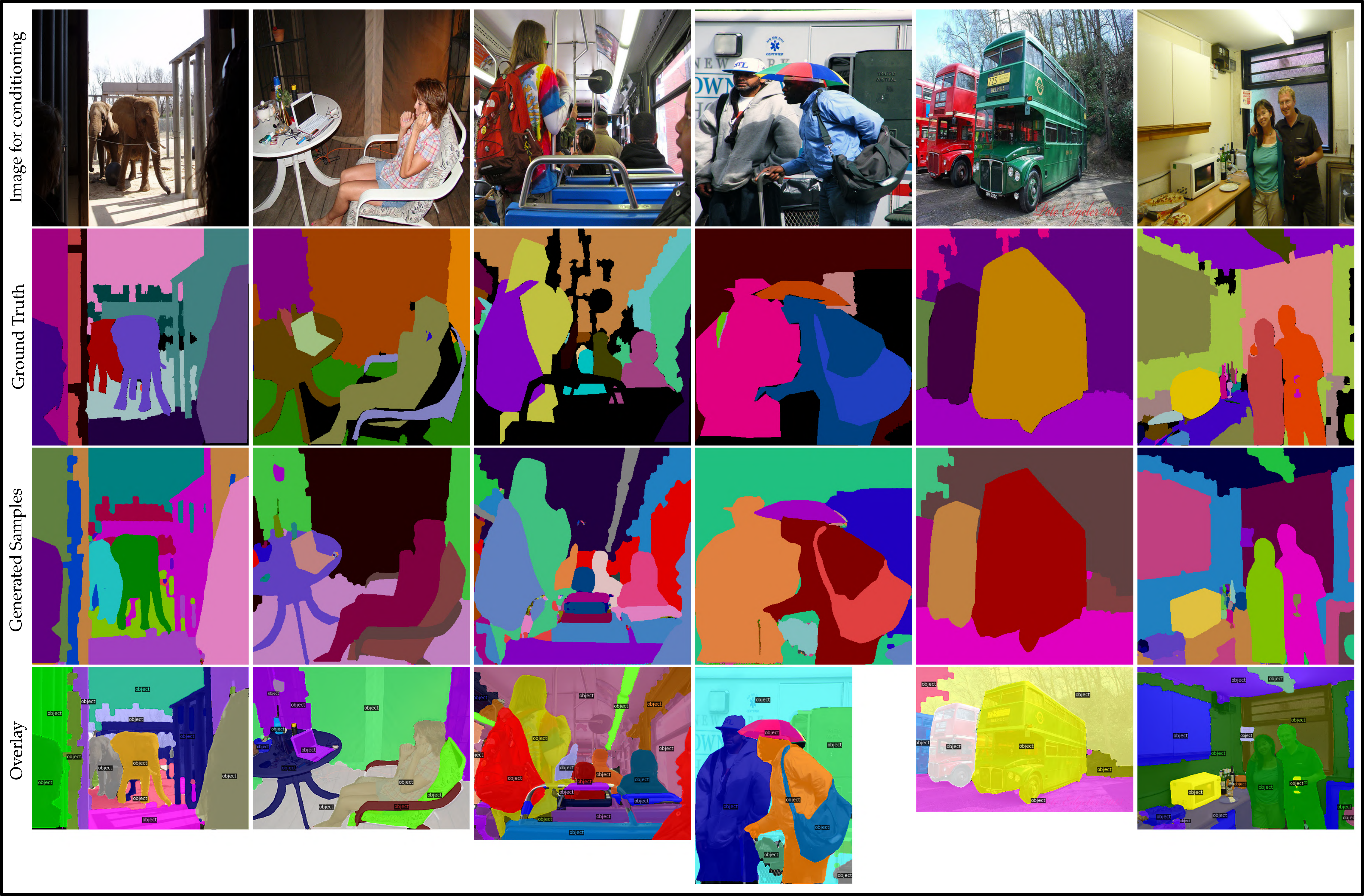}
   \caption{\textbf{Examples on COCO (2). } The figure displays more
   generated masks on the COCO val set.}
   \label{fig: generation_v2}
\end{figure*}

\paragraph{More Segmentation Results.} We show the panoptic segmentation results with 50 timesteps on COCO \texttt{val2017}~\cite{lin2014microsoft}  in Figure~\ref{fig: pq_results}. Additionally, we show (class-agnostic) masks in Figures~\ref{fig: generation_v1} and~\ref{fig: generation_v2}. The input images are resized to $3 \times 512 \times 512$ during training and the diffusion process acts on latents of size $4 \times 64 \times 64$. To visualize the masks, we assign each segment to a random color. Overall, the model is capable of generating high-quality panoptic masks.

\begin{figure*}
  \centering
   \includegraphics[width=1.0\linewidth]{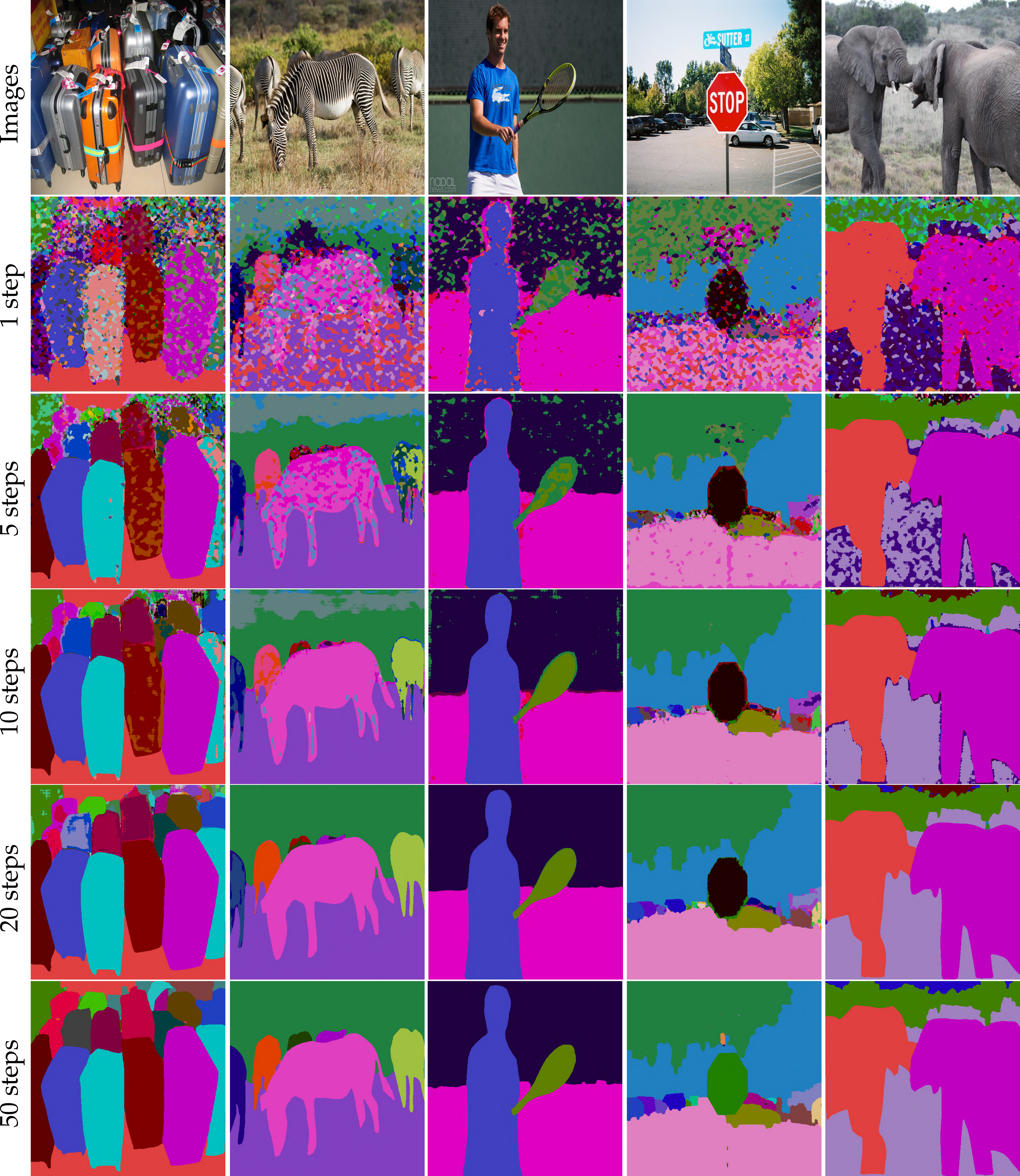}
   \caption{\textbf{Results for different timesteps. } The figure visualizes the image-conditioned samples for the timesteps 1, 5, 10, 20, and 50 in the diffusion process. Longer sampling is required to capture more details, which is beneficial for complex scenes (\eg, cars in the background in column 4).}
   \label{fig: results_timesteps}
\end{figure*}

\paragraph{Number of Denoising Steps.} Figure~\ref{fig: results_timesteps} displays the results for different timesteps during the denoising process. Longer sampling benefits the generation of details, such as capturing small objects in the background or an object's edges. This approach necessitates 10 - 50 iterations to produce high-quality segmentation masks, which is in line with latent diffusion models for images~\cite{rombach2022high}.
Furthermore, as the model was forced to distinguish between different instances during training, it's unlikely that different instances will be grouped during inference. Interestingly, the model iteratively improves the predictions while not reinforcing mistakes during the generative process.

\paragraph{Inference Time.} Table~\ref{tab: inference_time} provides the inference times for different sampling durations. In comparison, Painter requires approximately 0.5 and 0.7 seconds to post-process an image at a resolution of 448 and 560 respectively on our machine. Our post-processing method is significantly faster, taking up only about 0.024 seconds. Importantly, the performance will vary based on hardware and system specifications. Our relatively simple post-processing is explained in Supplement~\ref{sec: implementation_suppl} (final paragraph). 
Finally, recent research~\cite{song2023consistency} on Consistency Models looks promising to generate high-quality masks in a single step.

\paragraph{Encoding Panoptic Maps.} Table~\ref{tab: encoding} verifies our hypothesis w.r.t. the encoding scheme, as discussed in Sec. 3.1 (main paper). In particular, we test 3 encoding schemes: color (RGB) encoding \vs bit encoding \vs positional encoding: 
\begin{itemize}[]
    \item \textbf{Colors:} we generate 256 equidistant colors within the RGB space. 
    \item \textbf{Bits:} we employ 8 channels to represent integers from $[0, 255]$ using bits.
    \item \textbf{Positional:} we map integers from $[0, 255]$ to an 8-dimensional embedding following~\cite{mildenhall2020nerf}.
\end{itemize}
The mIoU and class-agnostic PQ are adopted to measure the reconstruction quality of the autoencoder. We hypothesize that the mapping from color to instance is sub-optimal as this scheme is sensitive to the chosen color palette (89.9 \vs 89.1\% PQ). In contrast, bit encoding is a general way to represent discrete panoptic maps, which also outperforms positional encoding (89.9 \vs 88.2\% PQ).

\paragraph{Tokenizers and Component Analysis.} Table~\ref{tab: component_analysis} shows that image tokenizers with more semantically meaningful image features can boost the results. In addition, we show the impact of employing different schedulers and an exponential moving average of the model weights. Note that the results are provided with 50 timesteps during inference. All components further enhance the performance of LDMSeg. To summarize, our best results are obtained with a ViT-B~\cite{dosovitskiy2020image} architecture and DINOv2~\cite{oquab2023dinov2} weights as the image encoder, the DDPM scheduler~\cite{ho2020denoising} and an exponential moving average of the model weights during training (weight of $0.999$).

\paragraph{Loss Weights.} Finally, we note that lowering the loss for small timesteps (\eg, $j < 25\%$) is not crucial, but speeds-up training by  0.3 to 0.5\% PQ. We aim to remove this in future work.

\begin{table}[H]
\caption{\textbf{Encoding.} Reconstruction quality for different encoding schemes.}
\tablestyle{10.0pt}{1.0}
    \begin{tabular}{l  c  c  c  c}
    \toprule
    \textbf{Encoding} & \textbf{mIoU} & \textbf{PQ [\%]} \\
    \midrule
    bit encoding & 97.3 &  89.9 \\
    color encoding & 97.0 & 89.1 \\
    positional encoding & 96.7 & 88.2 \\
    \bottomrule
    \end{tabular}
\label{tab: encoding}
\end{table}

\begin{table}
\caption{\textbf{Inference time.} We report the average time to generate a single panoptic mask on COCO with a 4090 GPU. The table provides the results for various denoising steps.} 
\resizebox{1.0\linewidth}{!}{
\tablestyle{8.0pt}{1.0}
    \begin{tabular}{c | c c c | c | c c c | c }
    \toprule
    & \multicolumn{3}{c|}{\textbf{Class-agn. Panoptic Seg.}} & \textbf{Sem. Seg.} & \multicolumn{3}{c|}{\textbf{Panoptic Seg.}} & \\
    \textbf{\# Iters} & \textbf{PQ [\%]} & \textbf{SQ [\%]} & \textbf{RQ [\%]} & \textbf{mIoU [\%]} & \textbf{PQ [\%]} & \textbf{SQ [\%]} & \textbf{RQ [\%]} & \textbf{Time [s]} \\
    \hline
        1 & 8.4	   &76.0	& 11.1		& 18.2		&8.1	&68.9	&10.8 & 0.115 \\
        2 & 35.5	&83.9	& 42.3		& 21.3		&19.8	&78.4	&24.8 & 0.160  \\
        3 & 42.4	&84.3	& 50.4		& 42.1		&35.5	&79.6	&43.8 & 0.207  \\
        4 & 45.5	&84.2	& 54.0		& 51.8		&39.3	&80.3	&48.2 & 0.259  \\
        5 & 47.3	&84.1	& 56.2		& 55.1		&41.3	&80.6	&50.3 & 0.320  \\
        \hline
        10& 50.2	&83.5	& 60.1		& 58.6		&43.4	&80.4	&52.6 & 0.575\\
        15&  51.0	&83.3	& 61.2		& 58.8		&43.7	&81.3	&53.0 & 0.815\\
        20& 51.4	&83.2	& 61.8		& 59.1		&44.1	&81.2	&53.4 & 1.071\\
        25& 51.7	&83.1	& 62.2		& 59.6		&44.3	&81.3	&53.7 & 1.336\\
        30& 51.8	&83.0	& 62.4		& 59.5		&44.1	&81.0	&53.7 & 1.585\\
        \hline
        40& 52.0	&82.9	& 62.7		& 59.3		&44.3	&81.1	&53.8 & 2.062  \\
        50& 51.9	&82.9	& 62.6		& 59.9		&44.3	&81.1	&53.8 & 2.548 \\
        60& 52.2	&82.8	& 62.8		& 59.3		&44.4	&81.2	&53.7 & 3.074 \\
        70& 52.2	&82.7	& 63.1		& 59.4		&44.3	&81.1	&53.7 & 3.564 \\
        80& 52.2	&82.6	& 63.1		& 59.3		&44.3	&80.5	&53.8 & 4.024 \\
        90& 52.2	&82.6	& 63.1		& 59.5		&44.3	&80.5	&53.7 & 4.550 \\
        \hline
        100& 52.1	&82.7	& 63.1		& 59.1		&44.3	&81.2	&53.7 & 5.030 \\
        200& 52.1	&82.5	& 63.2		& 59.1		&44.3	&80.5	&53.7 & 10.050\\

    \bottomrule
    \end{tabular}
}
\vspace{-0.2in}
\label{tab: inference_time}
\end{table}

\definecolor{Highlight}{HTML}{39b54a}  
\renewcommand{\hl}[1]{\textcolor{Highlight}{#1}}
\newcommand{\rownumber}[1]{\textbf{\textcolor{rowcolor}{#1}}}
\begin{table}
\caption{\textbf{Component Analysis.}}
\tablestyle{10.0pt}{1.0}
    \begin{tabular}{c  c  c  c  c}
    \toprule
    \textbf{~~Setup} & \textbf{Image Encoder} & \textbf{Scheduler} & \textbf{EMA} & \textbf{PQ [\%]} \\
    \midrule
    \rownumber{1} & SD VAE~\cite{rombach2022high}  & DDIM~\cite{song2021denoising} & \xmark & 40.3 \\
    \rownumber{2} & SD VAE~\cite{rombach2022high} & DDIM~\cite{song2021denoising} & \checkmark & 40.6 \\
    \rownumber{3} & ViT-B/14~\cite{dosovitskiy2020image} & DDIM~\cite{song2021denoising} & \checkmark & 43.7 \\
    \rownumber{4} & ViT-B/14~\cite{dosovitskiy2020image} & DDPM~\cite{ho2020denoising} & \checkmark & 44.3 \\
    \bottomrule
    \end{tabular}
\label{tab: component_analysis}
\end{table}

\section{Limitations and Future Work}
\label{sec: limitations}
Undoubtedly, our model has several limitations despite its general design.  
We discuss two limitations: (\textit{i}) the model can miss small background objects due to the projection to latent space; (\textit{ii}) the model is slower during inference than specialized segmentation models due to the adoption of a diffusion prior. 
In exchange, our method is simple, general and unlocks out-of-the-box mask inpainting. Moreover, the approach can be extended to a multi-task setting.
As we rely on plain diffusion models, new innovations (\eg, architectural, noise scheduler, tokenization, number of inference steps \etc.) in image generation are directly applicable to the presented framework.
Finally, increasing the dataset's size, increasing the latents' resolution, enabling open-vocabulary~\cite{radford2021learning} detection, and including more dense prediction tasks are exciting directions to explore further. 

\section{Broader Impact}
\label{sec: broader_impact}
The presented approach relies on pretrained weights from Stable Diffusion~\cite{rombach2022high}. Consequently, our model is subject to the same dataset and architectural biases. The user should be aware of these biases and their impact on the generated masks. For instance, these types of (foundation) models can hallucinate content.

\bibliographystyle{splncs04}
\bibliography{main}
\end{document}